\documentclass[letterpaper]{article} % DO NOT CHANGE THIS
\usepackage{aaai2026}  % DO NOT CHANGE THIS
\usepackage{times}  % DO NOT CHANGE THIS
\usepackage{helvet}  % DO NOT CHANGE THIS
\usepackage{courier}  % DO NOT CHANGE THIS
\usepackage[hyphens]{url}  % DO NOT CHANGE THIS
\usepackage{graphicx} % DO NOT CHANGE THIS
\usepackage{natbib}  % DO NOT CHANGE THIS AND DO NOT ADD ANY OPTIONS TO IT
\usepackage{caption} % DO NOT CHANGE THIS AND DO NOT ADD ANY OPTIONS TO IT
\usepackage{algorithm}
\usepackage{algorithmic}

\usepackage{newfloat}
\usepackage{listings}
\DeclareCaptionStyle{ruled}{labelfont=normalfont,labelsep=colon,strut=off} % DO NOT CHANGE THIS
\floatstyle{ruled}
\newfloat{listing}{tb}{lst}{}
\floatname{listing}{Listing}
\title{Multimodal DeepResearcher: Generating Text-Chart Interleaved Reports From Scratch with Agentic Framework}
\author{
    Zhaorui Yang\textsuperscript{\rm 1}{\equalcontrib},
    Bo Pan\textsuperscript{\rm 1}{\equalcontrib},
    Han Wang\textsuperscript{\rm 1}{\equalcontrib},
    Yiyao Wang\textsuperscript{\rm 1},
    Xingyu Liu\textsuperscript{\rm 1},
    Luoxuan Weng\textsuperscript{\rm 1}\\
    Yingchaojie Feng\textsuperscript{\rm 2},
    Haozhe Feng\textsuperscript{\rm 3},
    Minfeng Zhu\textsuperscript{\rm 4}\thanks{Corresponding Authors.},
    Bo Zhang\textsuperscript{\rm 4}\footnotemark[2],
    Wei Chen\textsuperscript{\rm 1}\footnotemark[2]
}
\affiliations{
    \textsuperscript{\rm 1}State Key Lab of CAD\&CG, Zhejiang University\\
    \textsuperscript{\rm 2}National University of Singapore\\
    \textsuperscript{\rm 3}Tencent TEG\\
    \textsuperscript{\rm 4}Zhejiang University
}

\usepackage[most]{tcolorbox}
\tcbuselibrary{breakable}
\usepackage{xcolor}
\usepackage{circledsteps}
\usepackage{booktabs}
\usepackage{multirow} 
\usepackage{multicol}
\usepackage{pdfpages}
\usepackage{marvosym}
\begin{document}

\maketitle

\begin{figure*}[t]
    \centering
    \includegraphics[width=\textwidth]{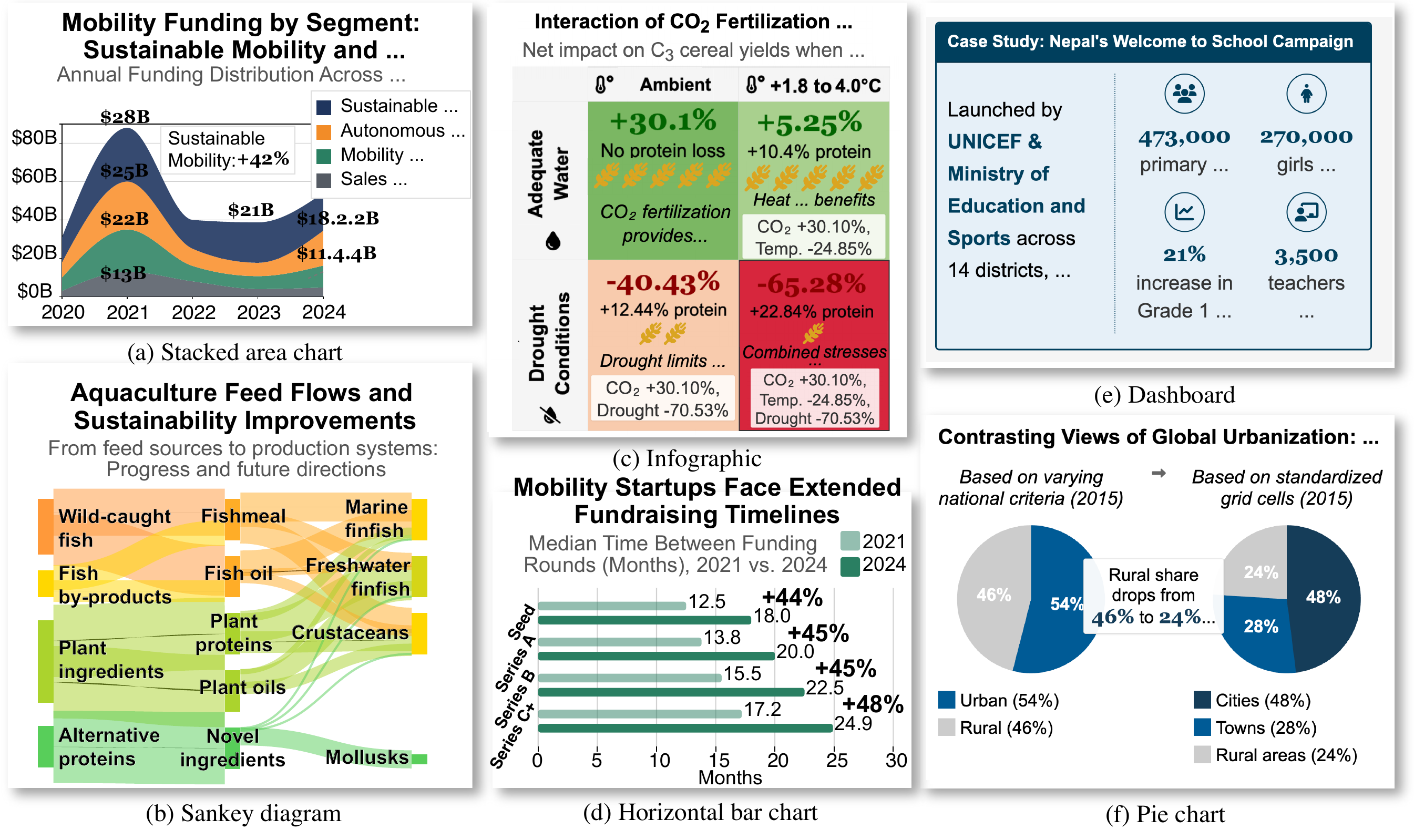}
    \caption{Examples of \textbf{visualization charts} generated by Multimodal DeepResearcher. \textbf{Accompanying texts are omitted} for brevity. As shown in the figure, it can produce diverse, high-quality charts beyond basic charts (e.g., line or bar chart).}
    \label{fig:case}
\end{figure*}

\begin{abstract}
Visualizations play a crucial part in effective communication of concepts and information. Recent advances in reasoning and retrieval augmented generation have enabled Large Language Models (LLMs) to perform deep research and generate comprehensive reports. Despite its progress, existing deep research frameworks primarily focus on generating text-only content, leaving the automated generation of interleaved texts and visualizations underexplored. This novel task poses key challenges in designing informative visualizations and effectively integrating them with text reports. To address these challenges, we propose Formal Description of Visualization (FDV), a structured textual representation of charts that enables LLMs to learn from and generate diverse, high-quality visualizations. Building on this representation, we introduce Multimodal DeepResearcher, an agentic framework that decomposes the task into four stages: (1) researching, (2) exemplar report textualization, (3) planning and (4) multimodal report generation. For the evaluation of the generated reports, we develop MultimodalReportBench which contains 100 diverse topics as inputs, and a set of dedicated metrics for report and chart evaluation. Extensive experiments across models and evaluation methods demonstrate the effectiveness of Multimodal DeepResearcher. Notably, utilizing the same Claude 3.7 Sonnet model, Multimodal DeepResearcher achieves an 82\% overall win rate over the baseline method.
\end{abstract}

\begin{links}
    % \link{Code}{https://github.com/rickyang1114/multimodal-deepresearcher}
    % \link{Datasets}{https://aaai.org/example/datasets}
    % \link{Extended version}{https://arxiv.org/pdf/2506.02454}
\end{links}

\section{Introduction}
Large language models (LLMs) have demonstrated broad capabilities in solving diverse tasks such as question answering, coding and math~\cite{bai2022training,deepseekr1,opencoder}. 
Augmented with searching and reasoning capabilities~\cite{openagents,webgpt,search-o1}, LLMs can perform deep research and effectively leverage up-to-date external information beyond static parameters~\cite{search-o1}.
Recently, this paradigm has garnered significant attention with its remarkable efficacy in generating grounded, comprehensive reports from scratch~\cite{storm,agentsroom}.
However, existing deep research frameworks from both academia~\cite{searchr1, deepresearcher} and industry~\cite{openaideepresearch, googledeepresearch,grokdeepresearch,dhdeepresearch} predominantly focus on generating textual content, neglecting the display beyond text modality. The text-heavy nature of these reports impedes effective communication of concepts and information~\cite{theoremexplainagent,pptagent}, which limits their readability and practical utility.

\begin{figure*}[t]
    \centering
    \includegraphics[width=\textwidth]{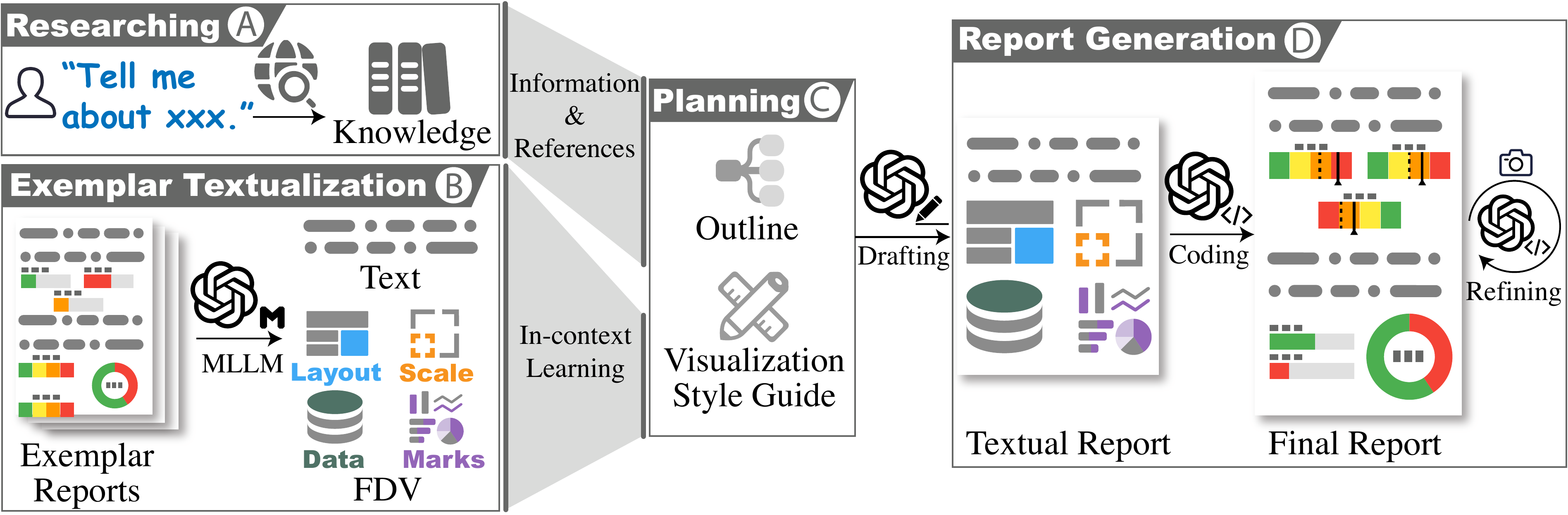}
    \caption{The framework of the Multimodal DeepResearcher. It decomposes the task of multimodal report generation into four stages: \Circled{A} Iterative researching about given topic; \Circled{B} Exemplar textualization of human experts using proposed Formal Description of Visualization (FDV); \Circled{C} Planning; \Circled{D} Report Generation, which generates the final report with crafting, coding and iterative refinement.}
    \label{fig:workflow}
\end{figure*}

In real-world scenarios, visualization serves as a crucial part of reports and presentations, offering remarkable capabilities for conveying data insights~\cite{otten2015infographics}, facilitating the identification of implicit patterns~\cite{matplotagent}, and enhancing audience engagement~\cite{barrick2018image,pptagent}.
Human experts typically craft meticulously designed visualizations with consistent styles to effectively communicate ideas and insights. They then integrate these visualizations within appropriate textual context~\cite{llm4vissurvey} to create coherent text-chart interleaved reports.

However, the end-to-end generation of multimodal reports remains challenging. Although LLMs are capable of generating individual charts through coding~\cite{matplotagent,vispath,chartllama}, effectively representing and integrating these visualizations with textual content still poses a challenge. While in-context learning appears to be a promising approach for guiding such generation, there lacks an appropriate representation to integrate text-chart interleaved content within the context of LLMs.

To address this challenge, we introduce the Formal Description of Visualization (FDV), a structured representation method inspired by the grammar of graphics~\cite{gog}, a classical visualization theory. FDV comprehensively captures visualization designs through four perspectives (i.e., overall layout, plotting scale, data, and marks). This representation provides universal and high-fidelity descriptions that enables in-context learning of multimodal reports from human experts, and can be generated to produce diverse and high-quality charts.

Building upon FDV, we introduce Multimodal DeepResearcher, an agentic framework that generates text-chart interleaved reports from scratch.  The framework operates through four stages: (1) researching, which gathers comprehensive information through searching and reasoning; (2) exemplar report textualization, which textualizes multimodal reports from human experts using our proposed Formal Description of Visualization (FDV) for in-context learning; (3) planning, which establishes a content outline and visualization style guide; and (4) multimodal report generation, which produces the final interleaved report through drafting, coding and iterative chart refinement. Some examples of the generated charts are presented in Figure~\ref{fig:case}.

We evaluate Multimodal DeepResearcher with MultimodalReportBench, which comprises 100 topics used as inputs. Our experiments include both proprietary and open-source models with automatic and human evaluation. The evaluation encompasses both report-level and chart-level assessments, each employing five dedicated metrics. As a baseline, we adapted DataNarrative~\cite{datanarrative}, a relevant framework that generates simple placeholders for charts from tabular inputs, to perform our task. Both automatic and human evaluations consistently demonstrate Multimodal DeepResearcher's superior performance compared to the baseline. Notably, when using Claude 3.7 Sonnet as the generator, Multimodal DeepResearcher achieves an impressive 82\% overall win rate.

Our contributions can be summarized as follows:

\begin{itemize}
    \item We introduce a novel task that generates a text-chart interleaved multimodal report from scratch and a corresponding dataset and evaluation metrics.
    \item We propose Formal Description of Visualization (FDV), a structured textual representation of visualizations that enables the in-context learning and generation of multimodal reports.
    \item We introduce Multimodal DeepResearcher, an end-to-end agentic framework that generates high-quality multimodal reports, which largely outperforms the baseline method.
\end{itemize}

\section{Related Work}
\paragraph{Deep Research}
Recently, the combination of retrieval techniques~\cite{li2025matching,zhao2024retrieval} and reasoning~\cite{deepseekr1} has enabled LLMs to transcend their parametric constraints by leveraging external knowledge. Pioneering works have designed specialized prompts and workflows for complex research tasks, as exemplified by OpenResearcher~\citep{openresearcher} and Search-o1 \citep{search-o1}. Subsequent research explored reinforcement learning for end-to-end reasoning and information retrieval~\cite{searchr1, deepresearcher}. However, these studies primarily focus on generating and evaluating text-only results, whereas this work advances the field by generating text-chart interleaved reports that significantly enhance information comprehension and communication with visualizations.

\paragraph{LLM for Data Visualizations}
Current work has focused on enhancing individual chart quality through various approaches, including multi-stage pipelines~\cite{lida}, iterative debugging with visual feedback~\cite{matplotagent}, chain-of-thought prompted query reformulation~\cite{vispath}, and models fine-tuned with domain-specific data for chart generation~\cite{chartllama, chartgpt}. Another line of work has explored how to articulate generation intent, such as multimodal prompting with sketches and direct manipulations~\cite{VisPilot}, multilingual natural language interfaces~\cite{Chat2VIS}, and conversational context management~\cite{ConversationalAI}. Corresponding evaluation methodologies have also been proposed~\cite{VisualizationAnEvaluation,viseval}. However, previous work has predominantly focuses on generating individual charts with limited data. To the best of our knowledge, we are the first to explore generating and evaluating text-chart interleaved reports with multiple visualizations, based on in-the-wild and heterogeneous information.

\paragraph{LLM for agentic generation}
LLMs have been widely applied to various generation tasks due to their ability to process complex textual information~\cite{ku-etal-2024-viescore,nijkamp2022codegen,nijkamp2023codegen2,jimenez2024swebench,yang2024swebenchmultimodal}. For challenging tasks that require multiple steps, researchers have designed LLM agents that decompose problems into reasoning, planning, and execution stages~\cite{luo2025largelanguagemodelagent}. These agents have demonstrated remarkable success across scientific research~\cite{lu2024aiscientist,si2024llmsgeneratenovelresearch,li2024mlrcopilotautonomousmachinelearning,bogin2024superevaluatingagentssetting}, video generation~\cite{he2025kubrickmultimodalagentcollaborations}, and computer system interaction~\cite{osworld,mind2web,yang2023appagent}. This paradigm extends effectively to the visualization domain as well. TheoremExplainAgent~\cite{theoremexplainagent} uses agents to generate educational videos, and PPTAgent~\cite{pptagent} automatically creates slides for presentation with integrated text and visuals. Most relevant to our work, DataNarrative~\cite{datanarrative} explores generating simple specifications for data-driven visualizations and evaluating these specifications as proxies for actual charts. However, this approach remains limited to simple chart types such as bar chart and line chart, which restricts its practical utility.

\section{Method}
\begin{figure*}[t!]
    \centering
    \includegraphics[width=\textwidth]{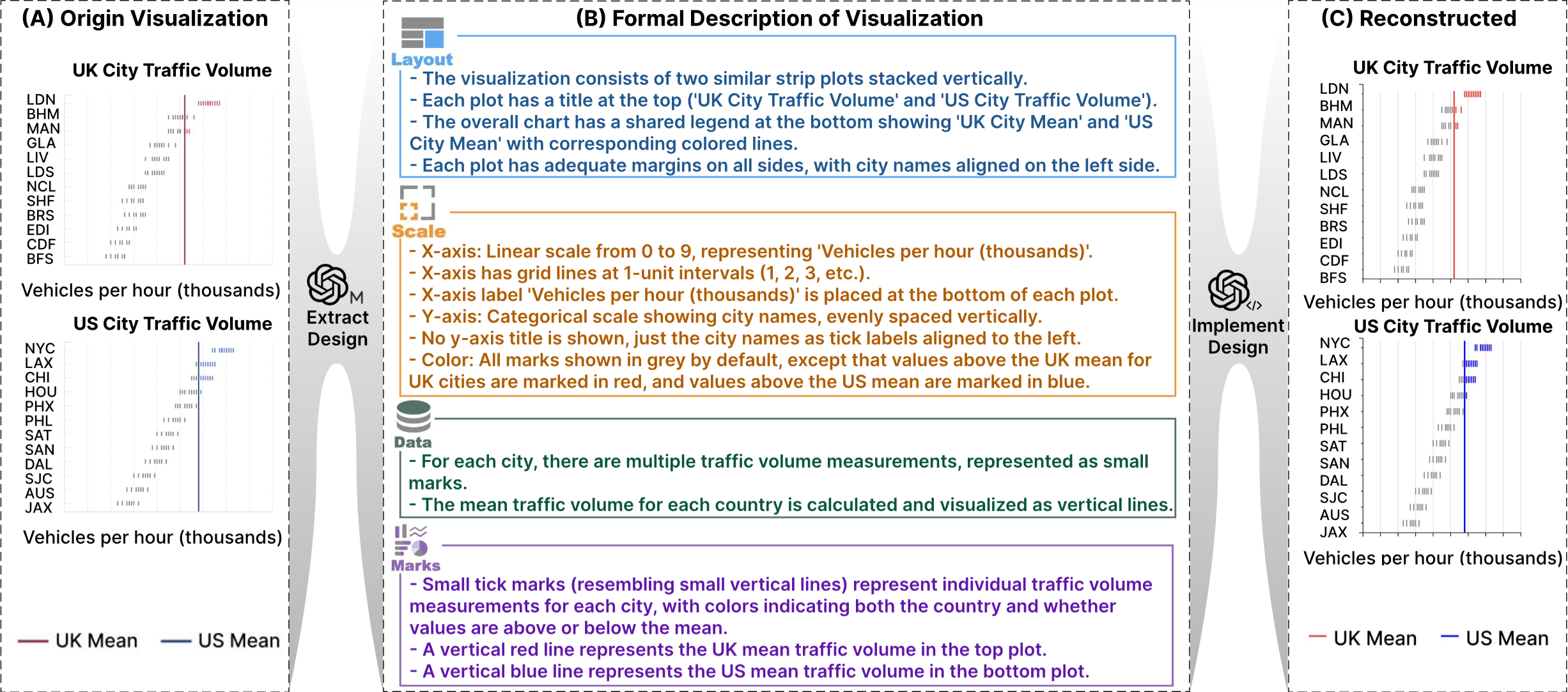}
    \caption{The illustration Formal Description of Visualization (FDV) for the exemplar textualization process. (A) Original traffic volume visualizations for UK and US cities; (B) The Formal Description of Visualization (FDV) that systematically captures the visualization's layout, scale, data, and marks using a structured format; and (C) The reconstructed visualization based on the formal description. This process textualizes high-quality text-chart interleaved reports by transforming visual elements into structured textual representations that preserve the visualization's essential characteristics.}
    \label{fig:FDV}
\end{figure*}

We formulate the task of multimodal report generation as follows: given a topic $t$ and a set of multimodal exemplar reports $R$ containing interleaved texts and charts, the system is expected to generate a multimodal report as in $R$ based on $t$. To solve this task, we introduce Multimodal DeepResearcher, an agentic framework which decomposes it into four steps: (1) researching through iterative web search and reasoning, (2) exemplar report textualization, which textualizes multimodal exemplar reports from human experts using proposed Formal Description of Visualization (FDV), (3) planning, and (4) Multimodal report generation. We present an overview of Multimodal DeepResearcher in Figure~\ref{fig:workflow}.

\subsection{Researching}
To leverage online information beyond parametric knowledge, Multimodal DeepResearcher conducts iterative research on a given topic $t$, generating a comprehensive set of learnings $L$. These learnings encompass both information acquired through web sources and their corresponding references. The process involves iterative execution of two primary operations: (1) web search and (2) subsequent reasoning based on search results. Initially, the agent prompts the LLM to generate relevant keywords $K = { k_1, \cdots, k_{n_K}}$ based on the given topic $t$. The agent then conducts web searches using these keywords and retrieves webpages $P = { p_1, \cdots, p_{n_P}}$. Subsequently, the agent analyzes these webpages, synthesizes the information into learnings $L$, and formulates a research question $q$ for the next iteration. Based on this research question and the original topic, the research agent performs the next research cycle. After $n_R$ rounds of iteration, a final list of learnings and references are produced. More details are provided in Appendix~\ref{appendix:details-researching}.

\subsection{Exemplar Textualization} 
Human experts typically produce reports with both texts and visualizations to enhance communication and audience engagement~\cite{pptagent,matplotagent}. To generate high-quality multimodal content comparable to expert-created reports, we employ in-context learning with exemplar reports crafted by human experts. This approach necessitates an effective methodology for converting multimodal exemplar reports $R$ into textual exemplar reports $\tilde{R}$.

To address this challenge, we propose Formal Description of Visualization (FDV), a structured description method for visualization charts inspired by the grammar of graphics (GoG) theory~\cite{gog}, which theoretically provides universal and high-fidelity descriptions for any visualization designs. As shown in Figure \ref{fig:FDV} (B), FDV characterizes each visualization chart from four perspectives: (1) Overall layout, detailing the constituent subplots and their spatial arrangements; (2) Plotting scale, describing the scaling logic behind each ``data to visual channel (e.g., position, color)'' mapping and their annotations; (3) Data, describing both the numeric data and text elements used to generate the visualization. (4) Marks, describing the design specifications of each visual element. The reverse process of textualization can be achieved via coding, which reconstructs the visualization from FDV, as shown in Figure~\ref{fig:FDV} (C). 

\begin{algorithm}[ht]
\caption{Textualization of multimodal reports}
\label{alg:fdv}
\begin{algorithmic}[1]

\STATE {\bfseries Inputs:} Multimodal exemplar reports $R$.
\STATE {\bfseries Requires:} Multimodal large language model $M_v$, replace function \textit{replace}.
\STATE {\bfseries Outputs:} Textualized exemplar reports $\tilde{R}$.
\STATE Initialize $\tilde{R} = \emptyset$
\FOR{$r$ {\bfseries in} $R$}
\STATE Init. $\tilde{r} = r$
\FOR{each image $i$ in $r$}
\STATE \textit{// Extract FDV from image}
\STATE $\text{FDV}_{i} = M_v(i)$
\STATE \textit{// Replace image with extracted FDV}
\STATE $\tilde{r} = \tilde{r}.\textit{replace}(i, \text{FDV}_i)$
\ENDFOR
\STATE $\tilde{R} = \tilde{R} \cup \{\tilde{r}\}$
\ENDFOR

\STATE {\bfseries Return: $\tilde{R}$} 

\end{algorithmic}
\end{algorithm}

In the exemplar textualization process, Multimodal DeepResearcher first extracts all visualization charts from the report, then prompts a multimodal large language model to extract the FDV representations of each chart. The FDV representations are then used to replace the charts. The algorithm for the process is presented in Algorithm~\ref{alg:refine}. Further details of FDV are provided in Appendix~\ref{appendix:details-textualization}, and the prompts are provided in Appendix~\ref{appendix:prompts-chartdesign}.

\subsection{Planning}
After iterative researching about the topic $t$, Multimodal DeepResearcher creates a plan before generating the final report. Specifically, it constructs an outline $O$ of the report to generate based on the learnings $L$, topic $t$ and textual exemplar report $\tilde{R}$. The outline comprises a hierarchical structure of sections, each with a descriptive title and a brief summary. To learn the style of visualizations in exemplar reports $\tilde{R}$ and maintain a consistent style of charts, Multimodal DeepResearcher also prompts the LLM to generate a visualization style guide $G$. The visualization style guide provides guidelines that control the overall style of visualizations in the report (e.g., color palette, font hierarchy). More details of this process can be found in Appendix~\ref{appendix:details-planning}.

\begin{algorithm}[t]
\caption{Algorithm for refining charts}
\label{alg:refine}
\begin{algorithmic}[1]

\STATE {\bfseries Inputs:} chart $c$ represented as code.
\STATE {\bfseries Requires:} Browser tool $T$, LLM $M_t$, Multimodal LLM $M_v$.
\STATE {\bfseries Outputs:} Refined chart $\tilde{c}$.
\STATE {\bfseries Hypars:} Number of max retry times $N_{max}$.
\STATE Initialize  $\text{satisfied} = \text{False}, c_0 = c$, $C = \{c\}$.
\FOR{$i = 1$ {\bfseries to} $N_{max}$}
\STATE \textit{// Get console message and image}
\STATE $\text{msg}, i = T(c)$
\STATE \textit{// Critic $M_v$ evaluates the chart}
\STATE $\text{satisfied, feedback} = M_v(i)$
\IF{satisfied == \text{True}}
\STATE break
\ENDIF
\STATE \textit{// actor $M_t$ refines previous chart}
\STATE $c_{i} = M_t(c_{i-1}, \text{msg}, \text{feedback})$
\STATE $C = C \cup \{ c_i \}$
\ENDFOR
\STATE $\tilde{c} = c_0$
\IF{$\vert C \vert > 1$}
\STATE \textit{// Selects from the last two charts}
\STATE $\tilde{c} = M_v(C[-1], C[-2])$
\ENDIF

\STATE {\bfseries Return: $\tilde{c}$}

\end{algorithmic}
\end{algorithm}

\subsection{Final Report Generation}
The final stage of Multimodal DeepResearcher is to generate the multimodal report with interleaved textual content and visualizations. The report is generated with outputs of previous stages, i.e., learnings $L$, exemplar textual reports $\tilde{R}$, outline $O$ and visualization style guide $G$. 

Multimodal DeepResearcher first prompts the LLM to generate a textual report with Formal Description of Visualization (FDV) as a placeholder for the underlying visualization chart to be generated. The format of this textual report is expected to be the same as those in textual exemplar reports used for in-context learning. Then, Multimodal DeepResearcher extracts all occurrences of FDVs, and prompts the LLM to implement the design via coding. Since visualizations represented by FDV have extensive flexibility, which may exceed the expressive capabilities of typical declarative visualization libraries \cite{Declarative} (e.g., matplotlib), we directed the LLMs to utilize D3.js, the most widely used imperative visualization programming to implement the target visualization designs.

To further improve the quality of visualizations generated, we include an actor-critic mechanism to revise and refine the code for generating the charts motivated by recent advancements of agents~\cite{matplotagent}. In this scenario, the actor is the LLM $M_t$ that generates code for chart, and feedback comes from both console and a critic model. 

Console feedback is collected using chrome developer tool provided as Python package. It first tries to load each visualization, collecting all console message with errors or warnings during loading. After all elements are loaded, it takes a screenshot to obtain the visualization chart rendered.

After getting the screenshot of each visualization chart, Multimodal DeepResearcher employs a multimodal LLM (MLLM) $M_v$ to serve as a critic, which provides visual feedback. The MLLM takes the chart rendered as input, examines its visual quality, and delivers corresponding feedback. It further determines whether the current chart needs improvement. If improvement is needed, the actor refines its code based on the feedback and console message. This iterative refinement continues until the critic is satisfied, or a predefined upper limit of 
retry times is reached, which we set as $3$ to avoid infinite refinement cycles. When the refinement process finishes, the critic selects the final chart from the last two candidates during refinement.

The refine process is detailed in Algorithm~\ref{alg:refine}. The prompts are provided in Appendix~\ref{appendix:prompts-chart}. A comprehensive \textit{full report} generated by Multimodal DeepResearcher is presented in Appendix~\ref{appendix:example-report}.

\section{Experiments}

\subsection{Data Selection}
To systematically evaluate the multimodal report generated by Multimodal DeepResearcher, we constructed MultimodalReportBench, a benchmark comprising 100 real-world topics curated from public websites that feature multimodal reports crafted by human experts, i.e., Pew Research~\cite{pewresearch}, Our World in Data~\cite{owid} and Open Knowledge Foundation~\cite{okf}. Pew Research informs the public about issues, attitudes and trends shaping the world through research report. Our World in Data presents empirical data and research on global development challenges through web publications. The Open Knowledge Foundation is dedicated to promoting open data and content across all domains, ensuring information accessibility. These sources contain exemplary multimodal reports, making their topics appropriate for our task.

The topics are then used as inputs for multimodal report generation. To ensure that our dataset applies to the real-world scenario, we meticulously curated topics spanning 10 categories, such as travel, energy and education. The distribution of topic categories is provided in Appendix~\ref{appendix:details-data}. We also collected 6 multimodal reports with no overlapping in topics to serve as exemplar reports for in-context learning.

\subsection{Baseline Selection}
Our task requires generating a multimodal report from scratch, which is infeasible with direct prompting or existing deep research frameworks. Most existing visualization generation works either focus on single-chart generation \cite{lida,matplotagent,chartgpt} or requires human interactions~\cite{fu2025dataweaver,li2024we,shao2025narrative}, which deviates from our setting of automated generation. Most similar to our work, DataNarrative~\cite{datanarrative} generates simple data-driven visualization specifications based on data tables as input, and evaluates the textual specification as a proxy of chart. We incorporate our researching module and adapt its framework accordingly to establish our baseline. For fair comparison, we utilize the learnings generated with our researching stage and plans instead of tables as the input. It then goes through generate-verify-refine process, consistent with the original framework. Since the original framework lacks mechanisms for transforming design specifications into actual charts, we extract all design specifications and generate corresponding visualizations using the same pipeline as Multimodal DeepResearcher does.

\subsection{Framework Implementation}
Multimodal DeepResearcher is an agentic framework with multiple stages. In this section, we describe the implementation details of each stage. In the researching stage, we perform web search and conduct reasoning with GPT-4o-mini~\cite{gpt4omini}. GPT-4o-mini is also utilized for planning. Claude 3.7 Sonnet~\cite{claude37} is utilized as the MLLM for the textualization of exemplar reports. The generation of the final multimodal report requires both a large language model to craft textual report, and a multimodal large language model to provide visual feedback for the chart. Our experiments encompasses two model configurations: (1) State-of-the-art proprietary models, with Claude 3.7 Sonnet serving as both the LLM and multimodal LLM. (2) Open-source models, specifically Qwen3-235B-A22B~\cite{qwen3} and Qwen2.5-VL-72B-Instruct~\cite{Qwen2.5-VL}. To ensure fair comparison, all the settings are consistent in both Multimodal DeepResearcher and the DataNarrative baseline where applicable. 

\subsection{Automatic Report Evaluation}
Given the multimodal nature of the outputs in our task, evaluation necessitates assessment of both texts and visualizations. To accomplish this, we conducted both report-level and chart-level evaluation to comprehensively assess the quality of all reports. For automatic report evaluation, we task the evaluator (i.e., GPT-4.1) with pairwise comparison of reports, generated from the save topic with both methods. Since report generation constitutes an open-ended, subjective task, reference-based metrics typically fail to align with human-perceived standards~\cite{g-eval}. Therefore, we established a comprehensive criteria incorporating both texts and visualizations in reports, which primarily consists of five metrics:

\textbf{Informativeness and Depth.} Evaluates whether the report delivers comprehensive, substantive and thorough information through both texts and accompany visualizations.

\textbf{Coherence and Organization.} Evaluates whether the report is well-organized, and whether the visualizations connect meaningfully to the text. \looseness=-1

\textbf{Verifiability.} Evaluates whether the information of the reports can be verified with citations. Apart from textual links to references, we also prompt the evaluator to check the annotation present in visualizations that may contain source information.

\textbf{Visualization Quality.} Evaluates the quality of visualization charts in the report, including visual clarity and textual labels and annotations.

\textbf{Visualization Consistency.} Evaluates whether the visualizations in the report maintain a consistent overall style. The style contains the color palettes, typography and information hierarchy in visualizations.

During evaluation, we provide the evaluator with the topic, learnings which contain both knowledge acquired through web search, references, and both reports. Specifically, we employ rubric scoring on a 1-5 scale with detailed guidelines. The scores are then compared to determine which method is better or they tie. To mitigate potential positional bias, we randomize the order of reports. The complete prompts for evaluation are provided at Appendix~\ref{appendix:prompts-eval}.

\textbf{Results.} As illustrated in Table~\ref{tab:main}, Multimodal DeepResearcher consistently outperforms DataNarrative across both proprietary and open-source model configurations. With Claude 3.7 Sonnet, it achieves an overall win rate of 82\%. Specifically, Multimodal DeepResearcher outperforms with a high win rate in Verifiability (86\%), Visualization Quality (80\%) and Visualization consistency (78\%). A similar pattern is observed with open-source models, where Multimodal DeepResearcher achieves a win rate of 55\%. Notably, the performance advantage is more pronounced with Claude 3.7 Sonnet than with open-source models. This gap arises as Multimodal DeepResearcher requires multifaced capabilities, including planning, writing, coding, and refinement. Therefore, Multimodal DeepResearcher benefits more from a stronger model, whereas DataNarrative's simpler architecture limits its capacity to leverage model improvements. The results demonstrate the efficacy of its in generating multimodal reports. We also presented the raw scores obtained and results with other evaluators in Appendix~\ref{appendix:more-exp}. 

\begin{table}[t]
\small
\centering
\begin{tabular}{lccc}\\ & \multicolumn{3}{c}{\textbf{Ours vs DataNarrative}} \\ \midrule
\textbf{Evaluation Metrics} & Ours Win & Ours Lose & Tie \\
\midrule
\multicolumn{4}{c}{w. \textit{Claude 3.7 Sonnet}} \\
Informativeness and Depth & \textbf{75\%}  & 25\% & 0\%\\ 
Coherence and Organization & \textbf{76\%} & 21\% & 3\%\\ 
Verifiability & \textbf{86\%} & 5\% & 9\%\\ 
Visualization Quality & \textbf{80\%} & 16\% & 4\%\\ 
Visualization Consistency & \textbf{78\%} & 17\% & 5\%\\ 
Overall & \textbf{82\%} & 16\% & 2\% \\
\midrule
\multicolumn{4}{c}{w. \textit{Qwen3-235B-A22B \& Qwen2.5-VL-72B-Instruct}} \\
Informativeness and Depth & \textbf{50\%}  & 50\% & 0\%\\ 
Coherence and Organization & 41\% & \textbf{51\%} & 8\%\\ 
Verifiability & \textbf{66\%} & 21\% & 13\%\\ 
Visualization Quality & \textbf{48\%} & 46\% & 6\%\\ 
Visualization Consistency & \textbf{52\%} & 42\% & 6\%\\ 
Overall & \textbf{55\%} & 40\% & 5\% \\

\bottomrule
\end{tabular}
\caption{Automatic evaluation results of the multimodal report: Multimodal DeepResearcher (Ours) vs. DataNarrative.
}
\label{tab:main}
\end{table}

\subsection{Human Evaluation}
For human evaluation, we utilized the same set of metrics as in automatic report evaluation. We selected a random subset of 20 topics for evaluation. Specifically, 5 annotators performed pairwise comparison of reports generated by both Multimodal DeepResearcher and DataNarrative with Claude 3.7 Sonnet. As with automatic evaluation, we randomized the order to avoid potential positional bias. Results are presented in Table~\ref{tab:human-eval}. Surprisingly, Multimodal DeepResearcher achieves an overall win rate of 100\%. Specifically, two annotators preferred all 20 reports generated by Multimodal DeepResearcher, one annotator preferred 19 out of 20, another annotator preferred 18, and the last annotator preferred 15. Comparing with the results given by GPT-4.1, the agreement between them is 80\%. These results further validate the effectiveness of Multimodal DeepResearcher. 

\begin{table}[t]
\small
\centering
\begin{tabular}{lccc}\\ \toprule
\textbf{Evaluation Metrics} & Ours Win & Ours Lose & Tie \\
\midrule
Informativeness and Depth & \textbf{100}\%  & 0\% & 0\%\\ 
Coherence and Organization & \textbf{95\%} & 0\% & 5\%\\ 
Verifiability & \textbf{100\%} & 0\% & 0\%\\ 
Visualization Quality & \textbf{75\%} & 20\% & 5\%\\ 
Visualization Consistency & \textbf{90\%} & 0\% & 10\%\\ 
Overall & \textbf{100\%} & 0\% & 0\% \\
\bottomrule
\end{tabular}
\caption{Human evaluation of the generated reports: Multimodal DeepResearcher (Ours) vs. DataNarrative.
}
\label{tab:human-eval}
\end{table}

\subsection{Chart Evaluation}

\begin{table}[t]
\centering
\begin{tabular}{lcc}\\ \toprule
\textbf{Evaluation Metrics} & Ours & DataNarrative \\
\midrule
\multicolumn{3}{c}{w. \textit{Claude 3.7 Sonnet}} \\
Readability & \textbf{8.97}  &8.52\\ 
Layout & \textbf{9.23} & 8.48\\ 
Aesthetics & \textbf{9.12} & 8.38\\ 
Data Faithfulness & \textbf{9.83} & 9.59\\ 
Goal Compliance & \textbf{9.75} & 9.24\\ 
\midrule
\multicolumn{3}{c}{w. \textit{Qwen3-235B-A22B \& Qwen2.5-VL-72B-Instruct}} \\
Readability & \textbf{7.05}  & 6.85\\ 
Layout & \textbf{6.70} & 6.40\\ 
Aesthetics & \textbf{7.22} & 6.74\\ 
Data Faithfulness & 7.93 & \textbf{7.99}\\ 
Goal Compliance & \textbf{7.17} & 6.94\\ 

\bottomrule
\end{tabular}
\caption{Evaluation of chart quality. The evaluator assigns a score between 1 to 10 for each metric, and the results are average across all reports.}
\label{tab:eval-chart}
\end{table}

To provide a more fine-grained evaluation of our framework, we further conducted assessments of individual charts to examine their quality and fidelity. Following established practices in data-driven visualization~\cite{lida,viseval}, which provided explainable evaluations for charts, we curated five metrics: (1) \textit{Readability}, (2) \textit{Layout}, (3) \textit{Aesthetics}, (4) \textit{Data Faithfulness} and (5)~\textit{Goal compliance}. For each chart, we employed the evaluator to score based on the chart along with its original design specification. We then average the scores of all charts within each report. As demonstrated in Table~\ref{tab:eval-chart}, Multimodal DeepResearcher consistently outperforms DataNarrative, with particularly notable improvements in layout and aesthetics.

\subsection{Ablation Studies}
To assess the efficacy of individual components of Multimodal DeepResearcher, we conducted ablation experiments on a random subset of 20 topics. Specifically, we compared 3 variants against Multimodal DeepResearcher: (1) w/o in-context learning from exemplar reports (2) w/o planning (3) w/o iterative refinement of charts. To ensure fair comparison, all other settings and hyperparameters remained consistent across variants. As shown in Table~\ref{tab:ablation_results}, removing any component results in significant performance degradation. Specifically, eliminating exemplar learning from human reports yields a 70\% lose rate, direct generation without planning leads to 85\%, and removing chart refinement process loses in 80\% cases. We further employed alternative evaluators to examine the effect of exemplar learning. The overall win rate for Multimodal DeepResearcher is 70\% when using GPT-5 as the evaluator, and 60\% with Gemini-2.5-Pro. These findings demonstrate the contribution of each component in Multimodal DeepResearcher.

\begin{table}[t]
\centering
\begin{tabular}{lccc}
\toprule
\textbf{Ablated Components}  & Lose & Win & Tie \\\midrule
- w/o Exemplar Learning                                                       & 70\%        & 20\%       & 10\%        \\
- w/o Planning                                                          & 85\%        & 15\%       & 0\%        \\
- w/o Refinement of charts & 80\%        & 20\%       & 0\%        \\
\bottomrule
\end{tabular}
\caption{Results of ablation studies across three different setups. We report the lose, win and tie rates for each setup against the complete Multimodal DeepResearcher. Claude 3.7 Sonnet serves as both the LLM and MLLM here.
}
\label{tab:ablation_results}
\end{table}

\section{Analysis}

\subsection{Visualization Analysis}
\begin{figure}[htb]
    \centering
    \includegraphics[width=\linewidth]{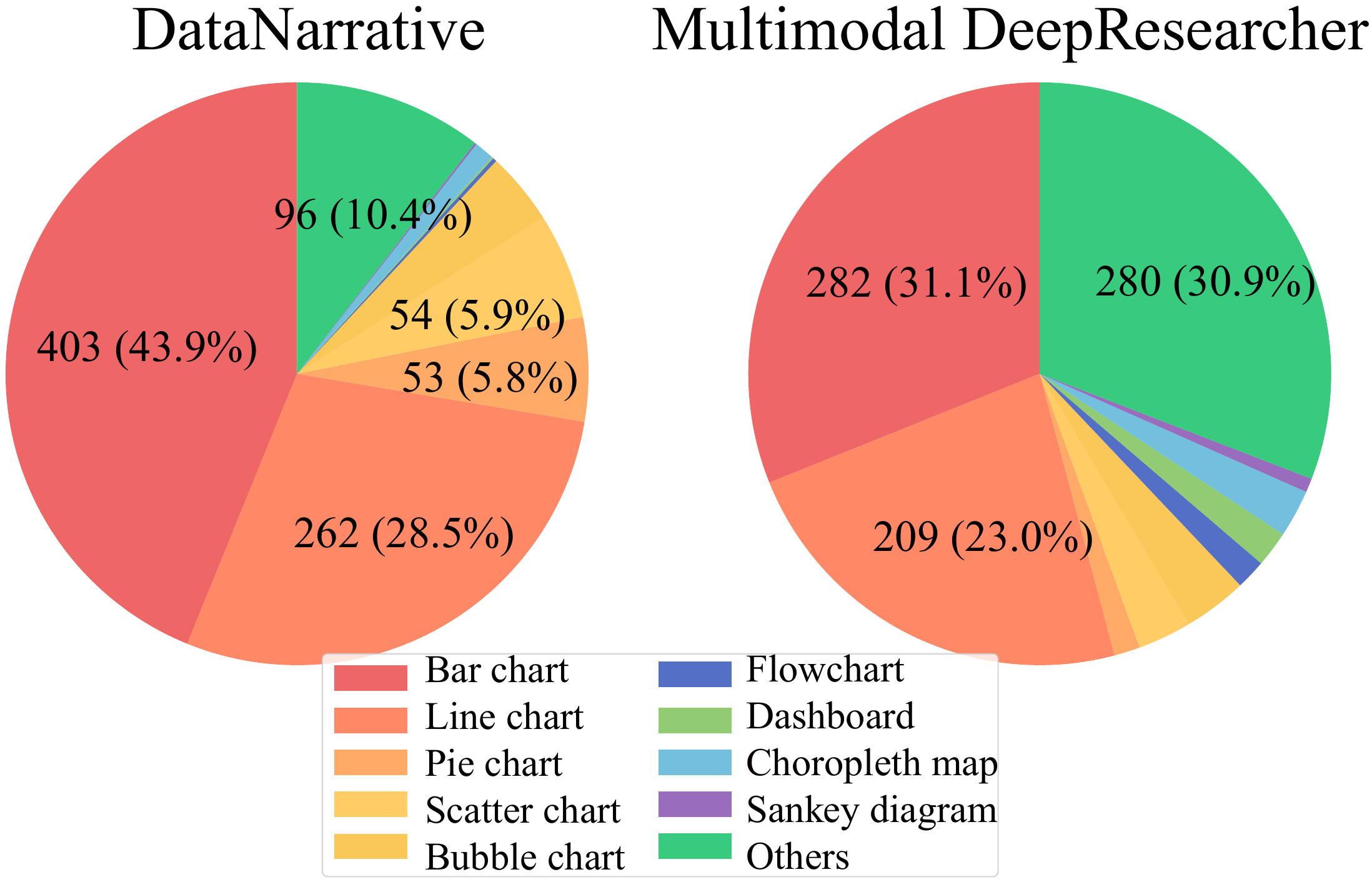}
    \caption{Distribution of visualization charts generated with DataNarrative and Multimodal DeepResearcher (Ours). The first column in the legend (denoted by red and yellow colors) represents conventional chart types.}
    \label{fig:chartsdistribution}
\end{figure}

In this section, we analyze the characteristics of visualizations generated with Multimodal DeepResearcher and the baseline. While the average number of charts per report between our framework (9.3) and DataNarrative (9.4) is comparable, the visualizations generated by Multimodal DeepResearcher are notably more diverse. As illustrated in Figure~\ref{fig:chartsdistribution}, although both methods prioritized basic chart types such as line chart and bar chart, Multimodal DeepResearcher demonstrates superior capability in generating sophisticated and complex visualizations.

For instance, across the 100 selected topics, Multimodal DeepResearcher produces 15 flowcharts and 18 dashboards, while DataNarrative generates merely 2 flowcharts and 1 dashboard. Another example involves the ``Others'' category, which encompasses hard-to-categorize visualizations such as infographics and mind maps. Our framework generates 280 such charts, substantially exceeding the 96 produced by DataNarrative. This disparity demonstrates that our approach can accommodate to diverse real-world scenarios. We provide a collection of examples for each type generated by our framework in Figure~\ref{fig:case} and Appendix~\ref{appendix:examples}.  

\subsection{Error Analysis}
Despite the remarkable efficacy of Multimodal DeepResearcher, the integration of visualizations poses new challenges. In this section, we categorize the identified common errors into the following two categories.
 
\paragraph{Overlapping} 
Element overlap represents the most prevalent error in the charts, primarily due to the inherent complexity of determining precise spatial positioning for all chart components without real-time visual feedback during the coding process. This error can be generally attributed to two factors: (1) excessive information in FDV that complicates proper arrangement within limited space. (2) suboptimal placement of legends, labels and annotations. Examples of both scenarios are provided in Appendix~\ref{appendix:examples-error}.

\paragraph{Hallucination}
Hallucination is a fundamental challenge for LLMs~\cite{storm}, which also extends to the generation of visualizations~\cite{datanarrative}. Despite explicit instructions to avoid creating fake data, models occasionally hallucinate when data is insufficient of unavailable. Figure~\ref{fig:map} in Appendix~\ref{appendix:choropleth} exemplifies this issue through a choropleth map chart. In this case, the model erroneously marked regions with inadequate data using red color, to denote the decline of a certain metric.

\subsection{Efficiency Analysis}
Another challenge for Multimodal DeepResearcher lies in balancing utility and efficiency. The system requires iterative refinement of multiple charts within reports. As demonstrated in our ablation study, this process significantly enhances the overall quality of generated reports. However, it also introduces computational overhead. In our experiments, we refine each chart for at most 3 iterations. After filtering out instances affected by network issues, the average generation time for a single report is 767.20 seconds, compared to DataNarrative's 372.94 seconds. Further analysis reveals that the refinement process accounts for the majority of execution time, requiring interaction with headless browsers, evaluation by multimodal large language models, and code regeneration. We plan to explore more precise critique mechanisms in future work.

\section{Conclusion}
In this work, we investigate the challenge of generating multimodal reports from scratch. We introduce the Formal Description of Visualization, a structured representation of charts that enables in-context learning from human-created exemplar reports. Based on this, we propose Multimodal DeepResearcher, an end-to-end framework for the generation of multimodal reports. While extensive experiments using both automatic and human evaluation confirm the efficacy of our framework, several challenges remain, including improving visualization quality, reducing hallucination, and balancing utility with efficiency.

\section*{Acknowledgments}
This work is supported by National Key R\&D Program of China under Grant No. 2024YFB4505500 \& 2024YFB4505503, National Natural Science Foundation of China (No. 62132017, No. 62421003 and No. 62402434), and  Zhejiang Provincial Natural Science Foundation of China (No. LD24F020011 and No. LQ24F020006). The work is partially conducted during Zhaorui Yang’s internship at the Machine Learning Platform Department, Tencent TEG. We thank Tencent Cloud BI for their support in the commercial implementation of the data reporting feature.
\bibliography{aaai2026}

% appendix begin
\clearpage
\appendix
\setcounter{secnumdepth}{2}
The appendices provide supplementary material organized as follows: implementation details of Multimodal DeepResearcher (Appendix~\ref{appendix:details}), additional experimental results and analysis (Appendix~\ref{appendix:more-exp}), prompts employed in Multimodal DeepResearcher and evaluation (Appendix~\ref{appendix:prompt}), representative examples of each chart type (Appendix~\ref{appendix:examples}), error case examples referenced in the main submission file (Appendix~\ref{appendix:examples-error}), a full report generated with Multimodal DeepResearcher (Appendix~\ref{appendix:example-report}).

\section{Implementation Details}
\label{appendix:details}
\subsection{Researching Details}
\label{appendix:details-researching}
In all of our experiments, model responses are sourced from OpenRouter platform with no GPU utilized to facilitate reproduction. Web search is implemented using Firecrawl API. In all of our experiments, we set the number of iterations $n_R$ to 2, the number of keywords generated $n_K$ to 3, the number of web pages retrieved for each keyword $n_P$ to 3, and the number of learnings generated from the research on each keyword $n_L$ to 3. We use the default hyperparameters when prompting LLMs and MLLMs.

Initially, large language model $M_t$ generates $n_K$ semantically distinct keywords ${k_1,...,k_{n_K}}$ and next research goal from the research topic given by user and prior learnings. Prior learnings is incorporated as contextual constraints to avoid redundancy and ensure exploratory diversity.

For each keyword $k_i$, the agent $M_t$ conducts web search to obtain $n_P$ webpage documents in Markdown format. The agent then filters duplicate contents through URL-based comparison and extract textual contents and semantic metadata from retained documents. The metadata is preserved as the reference. The agent analyzes documents and synthesizes it into $n_L$ learnings and $n_K$ questions as follow-up research directions for the next iteration. This step is guided by the prompt for learning generation in Appendix~\ref{appendix:prompts-keywords}.

After completing these two steps, the agent integrates the obtained next research goal and follow-up research directions to serve as the new topic for initiating the next round of search cycle. In the next iteration, $n_K$ is reduced by half and rounded up, thereby reflecting the gradual concentration of the search breadth as the search depth increases. After $n_R$ rounds of iteration, the researcher finally returns a list of final learnings and all the references. The workflow of the researching process is presented in Algorithm~\ref{alg:search}.

\begin{algorithm}[ht]
\caption{Algorithm for the research process}
\label{alg:search}
\begin{algorithmic}[1]

\STATE {\bfseries Inputs:} Topic $t$.
\STATE {\bfseries Requires:} Search engine $E$, large language model $M_t$.
\STATE {\bfseries Outputs:} Research Learnings $L$.
\STATE {\bfseries Hypars:} Number of iteration $n_R$, number of pages $n_P$ returned each search.
\STATE Initialize  $L= \emptyset,\text{q} = \emptyset, \text{goal} = \emptyset$;
\FOR{$i = 0$ {\bfseries to} $n_R - 1$}
\STATE Init. $P = \emptyset$
\STATE \textit{// Generate keywords and research goal}
\STATE $k_{1,\cdots, n_K}, \text{goal} = M_t(t,\text{q},L)$
\FOR{$k$ {\bfseries in} $K$}
\STATE // \textit{Fetch result pages}
\STATE $\tilde{p}_{1, \cdots n_P} = E(k)$
\STATE $P = P \cup \{\tilde{p}_{1, \cdots, n_P}\}$
\ENDFOR
\STATE \textit{// Learn from page content to get learnings}
\STATE $\tilde{L} = M_t(P, \text{goal})$
\STATE \textit{// Generate question for next iteration}
\STATE q = $M_t(P,\tilde{L})$
\STATE $L = L \cup \tilde{L}$
\ENDFOR

\STATE {\bfseries Return: $L$}

\end{algorithmic}
\end{algorithm}

Notably, the keywords employed for retrieval in our system are organized and precise. As illustrated in Algorithm~\ref{alg:search}, keywords are generated base on topic, question, and previous learnings. Such adequate information enables the system to generate organized and detailed keywords, substantially reducing ambiguity. To illustrate more concretely, we present a case study examining three representative keywords generated for the topic ``Air conditioning causes around 3\% of greenhouse gas emissions. How will this change in the future?'' during the initial iteration:

\begin{itemize}
    \item Global projections of greenhouse gas emissions from air conditioning through 2050
    \item Future cooling energy demand growth and its impact on AC-related CO2 emissions
    \item Emerging air-conditioning technologies and their potential to reduce future greenhouse gas output
\end{itemize}

Furthermore, each iteration incorporates information from multiple sources, thereby enabling cross validation that inherently filters out off-topic or low-quality information. This approach aligns with methodologies employed in recent deep-research works, such as DeepResearcher~\cite{deepresearcher} and WebThinker~\cite{webthinker}.

\subsection{Exemplar Textualization Details}
\label{appendix:details-textualization}

To provide a unified representation to describe visualization designs such that LLM can use it to effectively textualize existing visualization designs, we propose Formal Visualization Description (FDV) which is elaborated as follows.  

While GoG (grammar of graphics) \cite{gog} provide a rigorous mathematical model to describe visualizations, directly applying it has two key limitations for our usage scenario. First, there are still many visualization design choices can hardly be described by mathematical language. Second, using pure mathematical representation miss the opportunity to leverage the intuitiveness of natural language and substantial visualization design knowledge representation in natural language, which LLMs' has heavily seen during their training phases.

Thus, we design FDV as a practical extension of GoG that adheres to GoG's core framework while incorporating the strengths of natural language for describing visualization designs. Specifically, each component of FDV is backend by a formal definition extended from GoG, while specific design choices made in each component are expressed in natural language. 

Here we provide a formal definition for FDV: FDV describes the design choices (i.e. $\mathcal{F}_{\text{data}}$, $\mathcal{F}_{\text{mark}}$, $\mathcal{F}_{\text{scale}}$, $\mathcal{F}_{\text{layout}}$) made during the pipeline of transforming raw data into visualization as follows:

$$D_{\text{raw}} \xrightarrow{\mathcal{F}_{\text{data}}} D_{\text{plot}} \xrightarrow{\mathcal{F}_{\text{mark}} \& \mathcal{F}_{\text{scale}}} S \xrightarrow{\mathcal{F}_{\text{layout}}} V$$

where:
\begin{itemize}
    \item $D_{\text{raw}}$: Raw input data
    \item $D_{\text{plot}}$: Processed data for plotting
    \item $S = \{s_1, s_2, \ldots, s_n\}$: Set of rendered subplots
    \item $V$: Final visualization
\end{itemize}

$\mathcal{F}_{\text{data}}$: Describes how to compute data for plotting based on the raw input data (using operations like mapping, filtering, aggregation, ...)

$\mathcal{F}_{\text{mark}}$: Describes mark types and channel-field bindings. Specifically, for each subplot $i$ and mark $j$, describes:
$$\text{Mark}_{i,j} = (\text{type}, \text{encoding})$$

\begin{itemize}
    \item $\text{type} \in \{\text{point},\, \text{line},\, \text{bar},\, \text{area},\, \text{customized mark},\, \ldots\}$
    \item $\text{encoding} = \{\,(\text{channel},\, \text{field}) \mid \text{channel} \in \mathcal{C},\, \text{field} \in D_{\text{plot}}\,\}$
    \item $\mathcal{C} = \{x,\, y,\, \text{color},\, \text{size},\, \text{shape},\, \text{opacity},\, \ldots\}$\ (visual channels)
\end{itemize}

$\mathcal{F}_{\text{scale}}$: Define data-to-visual mappings and annotations, specifically, for each subplot $i$ and scale $k$, describes:
$$\text{Scale}_{i,k} = (\text{domain}, \text{range}, \text{transform}, \text{guide})$$

\begin{itemize}
    \item $\text{visual channel}$: The visual channel for applying scaling
    \item $\text{domain}$: Input data value range
    \item $\text{range}$: Output visual parameter range
    \item $\text{transform}$: Mapping function (linear, log, categorical, $\ldots$)
    \item $\text{guide}$: Visual annotation for understanding the scaling (axis, legend, colorbar, $\ldots$)
\end{itemize}

$\mathcal{F}_{\text{layout}}$: Describes how to arrange subplots into final visualization.

where each subplot $s_i$ is composed of:
\begin{itemize}
    \item mark specifications: $\{\text{Mark}_{i,j} \mid j = 1, \ldots, m_i\}$
    \item scale mappings: $\{\text{Scale}_{i,k} \mid k = 1, \ldots, p_i\}$
    \item data for plotting: $D_{\text{plot}}^{(i)} \subseteq D_{\text{plot}}$
\end{itemize}

The rendering process combines these components:
$$s_i = \text{Render}(\{\text{Mark}_{i,j}\}, \{\text{Scale}_{i,k}\}, D_{\text{plot}}^{(i)})$$

\subsection{Planning Details}
\label{appendix:details-planning}

In the planning phase, we employ the prompts in~\ref{appendix:prompts-outline} to generate a structured outline $O$ and a visualization style guide $G$ based on the topic $t$, learnings $L$ and high-quality exemplar reports $\tilde{R}$. We have set comprehensive and detailed requirements for the generation of the outline, including the number of sections, the clarity of key points, the minimization of conceptual overlap between sections, and the overall coherence of the report. We have also specified the format for each section.

In addition to the outline, we also generate a visualization style guide to ensure consistency while accommodating different concepts. We instruct the agent to use color coding and information hierarchy of professional industry reports that resembles the style of exemplar reports. With the help of the exemplar reports appended at the end of the prompt, the agent is able to generate higher-quality outlines and visualization style guides, thereby laying a solid foundation for subsequent report generation.

\subsection{Data Details}
\label{appendix:details-data}

We have meticulously selected 100 topics from Pew Research~\cite{pewresearch}, Our World in Data~\cite{owid}, and the Open Knowledge Foundation~\cite{okf} to serve as inputs for the Multimodal DeepResearcher. These topics cover 10 different categories, including technology, population, education, travel, energy, etc. Investigating these topics holds great significance for addressing real-world problems. The distribution of topic categories is shown in Table~\ref{tab:topic_distribution}.

\begin{table}[ht]
\centering
\begin{tabular}{lc}
\toprule
\textbf{Topic Categories}  & \textbf{Count}  \\\midrule
Technology \& Media & 15 \\
Agriculture \& Food & 13 \\
Travel & 4 \\
Population & 8 \\
Healthcare & 15 \\
Public Sector & 3 \\
Energy & 9 \\
Climate \& Environment & 14 \\
Education & 6 \\
Economy \& Work & 13 \\
\bottomrule
\end{tabular}
\caption{The distribution of topic categories.}
\label{tab:topic_distribution}
\end{table}

\subsection{Evaluation Details}
\textbf{Report evaluation.} Our report evaluation consists of the following five metrics:

\textit{Informativeness and Depth.} Evaluates whether the report delivers comprehensive, substantive and thorough information through both texts and accompany visualizations.

\textit{Coherence and Organization.} Evaluates whether the report is well-organized, and whether the visualizations connect meaningfully to the text. \looseness=-1

\textit{Verifiability.} Evaluates whether the information of the reports can be verified with citations. Apart from textual links to references, we also prompt the evaluator to check the annotation present in visualizations that may contain source information.

\textit{Visualization Quality.} Evaluates the quality of visualization charts in the report, including visual clarity and textual labels and annotations.

\textit{Visualization Consistency.} Evaluates whether the visualizations in the report maintain a consistent overall style. The style contains the color palettes, typography and information hierarchy in visualizations.

The exact prompts for automatic report can be bound at Appendix~\ref{appendix:prompts-eval}.

\textbf{Chart evaluation.} For chart evaluation, we utilize the following five metrics:

\textit{Readability}: Is the chart easy to read with appropriate titles, labels and colors?

\textit{Layout}: Is the layout of the chart appropriate with few or none issues such as overlapping?

\textit{Aesthetics}: Are the aesthetics of the visualization appropriate and effective for the visualization type and the data?

\textit{Data Faithfulness}: Is the data in the chart faithful to the data provided design specification?

\textit{Goal compliance}: How well the chart meets the specified visualization goals?

The exact prompts for automatic chart can be bound at Appendix~\ref{appendix:prompts-chart-eval}.

\section{More Experiments and Analysis}
\label{appendix:more-exp}
While reporting win/loss is an intuitive way to display how our framework outperforms baseline, We further validate the effectiveness of our framework with the raw scores (from 1 to 5), with GPT-4.1 as the evaluator. The results are presented in Table~\ref{tab:eval-report-raw}.

\begin{table}[t]
\centering
\begin{tabular}{lcc}\\ \toprule
\textbf{Evaluation Metrics} & Ours & DataNarrative \\
\midrule
\multicolumn{3}{c}{w. \textit{Claude 3.7 Sonnet}} \\
Informativeness and Depth & \textbf{4,79}  &4.36\\ 
Coherence and Organization & \textbf{4.75} & 4.35\\ 
Verifiability & \textbf{4.86} & 4.31\\ 
Visualization Quality & \textbf{4.78} & 4.33\\ 
Visualization Consistency & \textbf{4.79} & 4.32\\ 
\midrule
\multicolumn{3}{c}{w. \textit{Qwen3-235B-A22B \& Qwen2.5-VL-72B-Instruct}} \\
Readability & \textbf{4.34}  & 4.31\\ 
Layout & 4.24 & \textbf{4.28}\\ 
Aesthetics & \textbf{4.66} & 4.24\\ 
Data Faithfulness & \textbf{4.06} & 3.96\\ 
Goal Compliance & \textbf{4.15} & 4.03\\ 

\bottomrule
\end{tabular}
\caption{Raw scores of reports with GPT-4.1 as the evaluator.}
\label{tab:eval-report-raw}
\end{table}

Apart from GPT-4.1, we also utilized another state-of-the-art MLLM, gemini-2.5-pro, to serve as the evaluator. We omitted claude 3.7 to avoid potential bias from utilizing the same model for both generation and evaluation. Table~\ref{tab:main-gemini} presents the results. As with GPT-4.1, Multimodal DeepResearcher still consistently outperforms DataNarrative~\cite{datanarrative} by a large margin. The win rates consistently surpasses 80\%, reaching 90\% and 93\% overall win rates with both model suites. 

When using Claude-3.7-Sonnet for generation, the overall agreement between the gpt-4.1 and gemini-2.5-pro is 78\%. In the other case, the agreement is 60\%. In terms of the agreement with human evaluation, The agreement between GPT-4.1 and human evaluators is 80\%, and the agreement between gemini-2.5-pro and human evaluators is 90\%.

\begin{table}[ht]
\small
\centering
\begin{tabular}{lccc}\\ & \multicolumn{3}{c}{\textbf{Ours vs DataNarrative}} \\ \midrule
\textbf{Evaluation Metrics} & Ours Win & Ours Lose & Tie \\
\midrule
\multicolumn{4}{c}{w. \textit{Claude 3.7 Sonnet}} \\
Informativeness and Depth & \textbf{82\%}  & 9\% & 9\%\\ 
Coherence and Organization & \textbf{86\%} & 12\% & 2\%\\ 
Verifiability & \textbf{92\%} & 1\% & 7\%\\ 
Visualization Quality & \textbf{87\%} & 12\% & 1\%\\ 
Visualization Consistency & \textbf{82\%} & 16\% & 2\%\\ 
Overall & \textbf{90\%} & 9\% & 1\% \\
\midrule
\multicolumn{4}{c}{w. \textit{Qwen3-235B-A22B \& Qwen2.5-VL-72B-Instruct}} \\
Informativeness and Depth & \textbf{84\%}  & 11\% & 5\%\\ 
Coherence and Organization & \textbf{87}\% & 8\% & 5\%\\ 
Verifiability & \textbf{90\%} & 0\% & 10\%\\ 
Visualization Quality & \textbf{90\%} & 9\% & 1\%\\ 
Visualization Consistency & \textbf{82\%} & 10\% & 8\%\\ 
Overall & \textbf{93\%} & 6\% & 1\% \\

\bottomrule
\end{tabular}
\caption{Automatic evaluation of multimodal reports utilizing gemini-2.5-pro as the evaluator.}
\label{tab:main-gemini}
\end{table}

\section{Prompts}
In this section, we provide the detailed prompt for each component of our Multimodal DeepResearcher framework, as well as the prompts for evaluation.

\label{appendix:prompt}

\onecolumn
\subsection{Prompt for SERP Query and Learning Generation}
\label{appendix:prompts-keywords}
The first prompt below is used to guide the agent to generate keywords for web searches based on the topic provided by the user. The second prompt aims to guide the agent to extract relevant information from the Search Engine Results Page (SERP) and generate learnings.
\begin{tcolorbox}[breakable, colback=white, title={Prompt for SERP Query Generation}]
\textbf{System Prompt:}\\
You are an expert researcher. Follow these instructions when responding:\\
- You may be asked to research subjects that is after your knowledge cutoff, assume the user is right when presented with news.\\
- The user is a highly experienced analyst, no need to simplify it, be as detailed as possible and make sure your response is correct.\\
- Be highly organized.\\
- Suggest solutions that I didn't think about.\\
- Be proactive and anticipate my needs.\\
- Treat me as an expert in all subject matter.\\
- Mistakes erode my trust, so be accurate and thorough.\\
- Provide detailed explanations, I'm comfortable with lots of detail.\\
- Value good arguments over authorities, the source is irrelevant.\\
- Consider new technologies and contrarian ideas, not just the conventional wisdom.\\
- You may use high levels of speculation or prediction, just flag it for me.\\\\
\textbf{User Prompt:}\\
Given the following prompt from the user, generate a list of SERP queries to research the topic. Return a maximum of \{queries\_num\} queries, but feel free to return less if the original prompt is clear. \\
Make sure each query is unique and not similar to each other: \\
$<$prompt$>$\{query\}$<$/prompt$>$\\
Here are some learnings from previous research:\\
\{learning\_str\}
\end{tcolorbox}
\begin{tcolorbox}[breakable, colback=white, title={Prompt for Learning Generation}]
\textbf{User Prompt:}\\
Given the following contents from a SERP search for the query $<$query$>$\{query\}$<$/query$>$, generate a list of learnings from the contents.\\
Return a maximum of \{learning\_num\} learnings, but feel free to return less if the contents are clear. Make sure each learning is unique and not similar to each other. The learnings should be concise and to the point, as detailed and information dense as possible.\\ Please seamlessly incorporate references to external sources using Markdown hyperlinks.\\
Make sure to include any entities like people, places, companies, products, things, etc in the learnings, as well as any exact metrics, numbers, or dates. The learnings will be used to research the topic further.\\
Extract all meaningful data available in the contents, including any tables or lists, and explictly contain them in the learnings.\\
In addition, return a list of follow-up questions to research the topic further, max of \{question\_num\}.\\
$<$contents$>$
\{contents\}
$<$/contents$>$
\end{tcolorbox}

\subsection{Prompt for Chart Design Extraction}
\label{appendix:prompts-chartdesign}
\begin{tcolorbox}[breakable, colback=white, title={Prompt for extracting formal discription of visualization from image}]
\textbf{System prompt:}\\
You are a visualization design expert. You will be given a visualization image, and your task is to extract the design document from the image. The design document should include the overall layout, plotting scale, data transform, and marks used in the visualization. Your description should be detailed enough that someone could accurately recreate the visualization based solely on your specifications.\\
\\
\textbf{User prompt:}\\
Extract a comprehensive and precise visualization design specification from the given image. Capture all visual elements, data representations, and design choices with exact measurements, positions, and relationships. Ignore branding elements like company logos or trademarks.\\
\#\# Overall Format\\
The format of the design document must strictly follow the following format:\\
$<$visualization$>$\\
\{\\
"Part-A: Overall Layout": \{\\
"Part-A.1": "...",\\
"Part-A.2": "...",\\
...\\
\},\\
"Part-B: Plotting Scale": \{\\
"Part-B.1": "...",\\
"Part-B.2": "...",\\
...\\
\},\\
"Part-C: Data": \{\\
"Part-C.1": "...",\\
"Part-C.2": "...",\\
...\\
\},\\
"Part-D: Marks": \{\\
"Part-D.1": "...",\\
"Part-D.2": "...",\\
...\\
\}\\
\}\\
$<$visualization$>$\\
\\
\#\# Explanation for Each Part:\\
\#\#\# Part-A: Overall Layout\\
* Description of the overall figure dimensions, margins, and background\\
* If there are multiple subplots, also describe the detailed breakdown of main component layout and positioning.\\
* Description of title, subtitle, and caption placements with specific alignments\\
* Analysis of whitespace usage and component spacing hierarchies\\
\\
\#\#\# Part-B: Plotting Scale\\
Describe each scale used (such as x-axis scale, y-axis scale, color scale). Be specific in the position, formatting, size and shape.\\
\\
\#\#\# Part-C: Data\\
Comprehensive listing of **ALL** exact data represented in the visualization. This includes titles, subtitles, axis labels, legends, and any other text or numerical data.\\
\\
\#\#\# Part-D: Marks\\
* Complete specification of all primary visual marks (bars, lines, points) with exact sizes.\\
* Text label specifications (font, size, weight, positioning relative to marks)\\
* Interaction between marks including overlaps, nestings, or connections\\
* Annotations, highlights, or emphasis techniques\\
* Color usage patterns and semantic meanings\\
* Text alignment and spacing patterns\\
\end{tcolorbox}

\subsection{Prompt for Outline Generation}
\label{appendix:prompts-outline}
The following prompt generates a report outline based on the topic and the learnings extracted from deep research.
\begin{tcolorbox}[breakable, colback=white, title={Prompt for Outline Generation}]
\textbf{System Prompt:}\\
You an expert report-generation assistant specialized in creating professional documents that combine insightful analysis with diverse visualizations. Your purpose is to help users transform raw information into polished, presentation-ready reports.\\
Below are a list of professional reports for your reference.\\
\#\# Example Reports\\
\{list\_of\_example\_reports\}

\textbf{User Prompt:}\\
Using the provided topic and previous learnings, please create a structured outline for a comprehensive report. The outline should present a logical narrative flow that thoroughly explores the subject matter. Please do NOT include introduction or conclusion sections.\\
\#\# Input\\\\
**Topic**\\
\{topic\}\\
\\
**Previous learnings**\\
\{learning\_str\}\\\\
\#\# Requirements\\\\
The outline should feature:\\
* 4-6 distinct sections forming a cohesive narrative progression\\
* Clear identification of key insights and report points within each section\\
* Minimal conceptual overlap between sections, with each section addressing unique aspects\\
* A clear and logical flow of ideas, ensuring that section are connected rather than isolated\\
\\
\#\# Deliverable Format\\\\
For each section, please provide:\\\\
**Title:** A concise, engaging heading that captures the section's essence\\
**Summary:** A brief narrative (3-5 sentences) synthesizing the key points and insights\\
\#\# Visualization Style Guide\\\\
Before detailing individual sections, please provide a foundational style guide for visualizations that ensures consistency while accommodating different concepts, including:\\\\
* **Base Design Elements:** Color palatte for common concepts across charts. Use color coding and information hierarchy of professional industry reports that resembles the style of example reports\\
This style guide should offer flexible guidelines rather than rigid specifications, allowing each visualization to effectively represent its concept while maintaining overall visual cohesion.\\
\end{tcolorbox}

\subsection{Prompt for Report Generation}
The following prompt is used to generate a report. In the system prompt, the format of the visualization part in the report is elaborated, and the meaning of each part of the format is provided. The user prompt generates a report with a specified visualization format based on the topic, learnings, and the visualization style guide extracted from high-quality reports.
\label{appendix:prompts-report}
\begin{tcolorbox}[breakable, colback=white, title={Prompt for Report Generation}]
\textbf{System Prompt:}
You an expert report-generation assistant specialized in creating professional text-image interleaved documents that combine insightful analysis with diverse visualizations.
When visualization is needed, generate a comprehensive and precise visualization design specification. Include all visual elements, data representations, and design choices with exact measurements, positions, and relationships.\\

\#\# Visualization format\\
The format of the design document must strictly follow the following format:\\
$<$visualization$>$\\
\{\{\\
"Part-A: Overall Layout": \{\{\\
"Part-A.1": "...",\\
"Part-A.2": "...",\\
...\\
\}\},\\
"Part-B: Plotting Scale": \{\{\\
"Part-B.1": "...",\\
"Part-B.2": "...",\\
...\\
\}\},\\
"Part-C: Data": \{\{\\
"Part-C.1": "...",\\
"Part-C.2": "...",\\
...\\
\}\},\\
"Part-D: Marks": \{\{\\
"Part-D.1": "...",\\
"Part-D.2": "...",\\
...\\
\}\}\\
\}\}\\
$<$visualization$>$\\
\\
\#\# Explanation for Each Part:\\
\#\#\# Part-A: Overall Layout\\
* Description of the overall figure dimensions, margins, and background\\
* If there are multiple subplots, also describe the detailed breakdown of main component layout and positioning.\\
* Description of title, subtitle, and caption placements with specific alignments\\
* Analysis of whitespace usage and component spacing hierarchies\\
* Consider creating composite visualizations where appropriate (for example, combining line and bar charts within a single subplot to enhance data comparison and maximize visual space).\\
\\
\#\#\# Part-B: Plotting Scale\\
Describe each scale used (such as x-axis scale, y-axis scale, color scale). Be specific in the position, formatting, size and shape.\\
\\
\#\#\# Part-C: Data\\
* Comprehensive listing of **ALL** necessary data for visualization. **ALL** data should be present or can be derived from provided learnings. Do not create fake data or add placeholders.\\
* Appropriate texts, including titles, subtitles, axis labels, legends and moderate amount of annotations.\\
\\
\#\#\# Part-D: Marks\\
* Complete specification of all primary visual marks (bars, lines, points) with exact sizes.\\
* Text label specifications (font, size, weight, positioning relative to marks)\\
* Interaction between marks including overlaps, nestings, or connections\\
* Annotations, highlights, or emphasis techniques\\
* Color usage patterns and semantic meanings\\
* Text alignment and spacing patterns\\

Below are a list of professional reports for your reference. Follow the style, including the layout, infomation hierarchy, stress of the visualization designs in these reports.\\
\#\# Example Reports\\
\{list\_of\_example\_reports\}\\\\

\textbf{User Prompt:}\\
Please generate a detailed report with interleaved texts and visualization based on the topic, outline and previous learnings.\\
\#\# Input\\
\#\#\# Topic of the report\\
\{topic\}\\
\\
\#\#\# Outline for the report\\
\{outline\}\\
\\
\#\#\# Previous learnings\\
\{learning\_str\}\\
\\
\#\#\# Visualization Style Guide\\
\{visualization\_style\_guide\}\\
\\
\#\# Guidelines\\
- When referencing the knowledge provided, include a Markdown hyperlink at the appropriate position using the source URL provided\\
- Maintain a professional, academic tone throughout\\
- Use second-level (\#\#) headings for the section title, and third-level (\#\#\#) headings for subsections\\
- only utilize data available in the previous learnings part. Do not create fake data or add placeholders.\\ 
\end{tcolorbox}

\subsection{Prompt for Chart Generation and Improvement}
\label{appendix:prompts-chart}
Initially, the chart generation prompt generates the complete visualization code for the charts based on the visualization part of the report. Subsequently, the chart evaluation prompt renders the visualized charts, takes screenshots, and conducts an assessment, providing suggestions for modifications. The chart regeneration prompt then regenerates the charts based on the improvements. The chart selection prompt is employed to compare two sets of visualization code and select the implementation that better meets the design criteria.
\begin{tcolorbox}[breakable, colback=white, title={Prompt for Chart Generation}]
\textbf{System prompt:}\\
You are a HTML, D3.js V7 implementation expert who transforms visualization designs into working code. You write clean, efficient HTML and D3.js code to create data visualizations exactly as specified. You follow D3.js best practices, optimize for performance, and ensure responsive design across devices.\\\\
\textbf{User prompt:}\\
I need a professional HTML visualization to convey insight based on provided visualization design specification. Please implement with html and d3.js according to the specifications below.\\
**Visualization Design Specification**\\
\{chart\_design\}\\
\#\# Implementation Requirements\\
- Ensure the visualization is located at the center and there is no large empty space\\
- The top-level wrapper should have no box-shadow, no margin, and no visible borders\\
- Use icons from font-awesome with $<$i$>$ tag and corresponding class name when needed\\
- Highlight key numbers with larger font size, font-family: 'Georgia', and deeper colors\\
\\
IMPORTANT: Deliver your solution as a complete, self-contained HTML file enclosed in a code block starting with "```html" and ending with "```" to ensure I can extract it properly.\\
\end{tcolorbox}

\begin{tcolorbox}[breakable, colback=white, title={Prompt for Chart Evaluation and Improvement}]
\textbf{System prompt:}\\
You are a HTML, D3.js V7 implementation expert who transforms visualization designs into working code. You write clean, efficient HTML and D3.js code to create data visualizations exactly as specified. You follow D3.js best practices, optimize for performance, and ensure responsive design across devices.\\\\

\textbf{Chart evaluation prompt:}\\
Here is a screenshot of the page rendered by the HTML code, along with any console messages that may contain errors. Please examine the image thoroughly and report any problems you find.
Specifically check for these common rendering issues:\\
\\
1. Placeholder content: Does the image contain placeholder text (e.g., "Lorem ipsum", "Chart title", "Sample data") instead of actual content?\\
2. Excessive annotations: Are there too many annotations or labels that clutter the visualization?\\
3. Overlapping elements: Do any text labels, legends, data points or other elements overlap, making content unreadable?\\
4. Sizing problems: Is the visualization too small to be readable or too large for its container? Does it have appropriate dimensions?\\
5. Excessive margins: Are there large empty spaces around the visualization?\\
\\
\#\# Console Message\\
\{console\_message\}\\
\\
For each issue found, provide:\\
1. A clear description of the issue\\
2. The specific location in the image where it occurs\\
3. Relevent elements that cause the issue\\
\\
Focus on learning issues. If no issues are found, end your response with "No issues found."\\\\
\textbf{Chart regeneration prompt:}\\
Based on the above evaluation, please regenerate the complete HTML code with all necessary fixes implemented. Ensure the new code:\\
\\
1. Addresses all the issues you identified\\
2. Maintains the overall functionality and design intent\\
3. Is complete and ready to run without additional modifications\\
\\
Specifically:\\
1. Remove redundant or overlapping annotations that don't add critical information\\
2. Reposition remaining annotations to ensure clear visibility and logical placement\\
3. Adjust chart dimensions or add annotations to increase overall size and eliminate excessive margins\\
4. Reduce the size of specific elements to prevent overlapping between components\\
5. Expand container dimensions to fully display truncated content\\
\\
IMPORTANT: Deliver your solution as a complete, self-contained HTML file enclosed in a code block starting with "```html" and ending with "```" to ensure I can extract it properly.\\
\end{tcolorbox}

\begin{tcolorbox}[breakable, colback=white, title={Prompt for Chart Selection}]
\textbf{System prompt:}\\
You are an expert in data visualization design. Your task is to evaluate the provided images based on the given design specification and select the most appropriate one.\\\\
\textbf{User prompt:}\\
Here are a visualization design specification and two charts that implement the specification, please identify which one best meets the following criteria:\\
* Most closely matches the design specification requirements\\
* Offers optimal readability (e.g., has least isses regarding overlapping, elements are of appropriate size and margin)\\\\
\#\# Visualization Design Specification\\
\{chart\_design\}\\
\#\# Response Format\\
Return your response in the following format:\\
\\
$<$evaluation$>$\\
$[$Your evaluation of the charts$]$\\
$<$/evaluation$>$\\
\\
$<$selection$>$\\
$[$first or second$]$\\
$<$/selection$>$
\end{tcolorbox}

\subsection{Prompt for Multimodal Report Evaluation}
\label{appendix:prompts-eval}

The following prompt is used to compare the quality of the reports generated by baseline and our Multimodal DeepResearcher through multi-dimensional scoring. The scores are compared to determine which one wins or they tie.

\begin{tcolorbox}[breakable, colback=white, title={Prompt for Report Evaluation}]
\textbf{System prompt:}\\
You are an expert evaluator of AI-generated reports with advanced knowledge of data visualization and information analysis. Your role is to provide fair, impartial assessments of report quality based strictly on objective criteria.\\
\\
\#\# Evaluation Task\\
You will evaluate two AI-generated reports based on:\\
- The overarching topic\\
- Research learnings from internet searches that are used as source of information for the reports\\
\\
For each criterion below, assign a score from 1-5 (1=poor, 5=excellent) with half-point increments allowed (e.g., 3.5). Provide a concise, evidence-based justification for each score, highlighting specific examples that demonstrate meaningful distinctions in quality between the reports. Your evaluation should clearly articulate why one report receives a higher or lower score than another based on observable differences in content, structure, or analysis. Be cautious with extreme scores (1 and 5).
\\
\#\# Evaluation Criteria\\
\#\#\# Informativeness and Depth: Does the report deliver comprehensive, substantive and thorough information?\\
Score 1: Extremely superficial content with minimal information. Contains only basic facts without context or explanation.\\
Score 2: Limited content with some relevant information but significant gaps. Lacks necessary depth on key aspects.\\
Score 3: Adequate information covering main points with some supporting details, but missing opportunities for deeper analysis.\\
Score 4: Comprehensive information with substantive details, examples, and insights across most sections.\\
Score 5: Exceptionally thorough coverage with rich, nuanced details, expert-level insights, and well-contextualized information throughout.\\
\\
\#\#\# Coherence and Organization: Is the report well-organized with visualizations that connect meaningfully to the text?\\
Score 1: Disorganized; lacks logical structure and coherence. Visualizations appear random and unconnected to text.\\
Score 2: Basic structure present but with awkward transitions. Visualizations loosely connected to surrounding content.\\
Score 3: Clear overall organization with occasional flow issues. Visualizations generally support the text but integration could be improved.\\
Score 4: Well-structured with smooth transitions between sections. Visualizations meaningfully integrated with text content.\\
Score 5: Impeccable organization with seamless progression of sections. Visualizations perfectly complement and enhance textual narrative.\\
\\
\#\#\# Verifiability: Does the infomation of the reports can be verified with citations?\\
Score 1: Rarely supported with evidence; many claims are unsubstantiated\\
Score 2: Inconsistently verified; some claims are supported; evidence is occasionally provided\\
Score 3: Generally verified; claims are usually supported with evidence; however, there might be a few instances where verification is lacking\\
Score 4: Well-supported; claims are very well supported with credible evidence, and instances of unsupported claims are rare.\\
Score 5: Very well-supported; almost every claim is substantiated with credible evidence, showing a high level of thorough verification.\\
\\
\#\#\# Visualization Quality: Do the visualizations in the report have excellent quality?\\
Score 1: Poor visualizations that confuse rather than clarify. Inappropriate chart types, missing labels, or misleading representations.\\
Score 2: Basic visualizations with few annotations or explanations; functional issues (e.g., unclear axes, poor color choices) hinder interpretation.\\
Score 3: Adequate visualizations with labels and annotations that communicate data clearly but lack refinement or miss opportunities for improved insight.\\
Score 4: Well-executed visualizations with great visual appeal, clear labeling and annotations, and thoughtful design choices.\\
Score 5: Expert-level visualizations that reveal insights through masterful design, appropriate annotations, and careful attention to visual communication principles \\
\\
\#\#\# Visualization Consistency: Do the visualizations in the report maintain a consistent style?\\
Score 1: No visual consistency. Charts use different color palettes, conflicting typography, inconsistent information hierarchy, and varying design treatments (such as different border styles, background treatments, or legend placements).\\
Score 2: Minimal consistency with obvious style variations across visualizations. While some basic elements might align, there are clear discrepancies in color usage, information organization, axis formatting, or label treatments.\\
Score 3: Moderate consistency with a partially unified approach. Most visualizations share similar color schemes and basic formatting, but variations exist in how information hierarchy is presented, how emphasis is applied, or how supporting elements are styled.\\
Score 4: Strong consistency with cohesive design elements. Visualizations share a clear color system, consistent information hierarchy, and unified styling approach, with only minor variations that don't distract from the report's overall visual flow.\\
Score 5: Perfect consistency across all visualizations with a meticulously applied design system. Unified color palette used purposefully to highlight key information, consistent information hierarchy that guides the viewer's attention appropriately, identical typography treatment, and harmonious spacing, scale, and proportion across all charts and graphics.\\
\\
\#\# Response Format:\\
Please give your response in the following XML format:\\
\\
$<$evaluation$>$\\
    $<$report\_a$>$\\
    $<$informativeness$>$\\
      $<$score$>$X$<$/score$>$\\
      $<$justification$>$\\
        Provide a brief justification here\\
      $<$/justification$>$\\
    $<$/informativeness$>$\\
    $<$coherence$>$\\
      $<$score$>$X$<$/score$>$\\
      $<$justification$>$\\
        Provide a brief justification here\\
      $<$/justification$>$\\
    $<$/coherence$>$\\
    $<$verifiability$>$\\
      $<$score$>$X$<$/score$>$\\
      $<$justification$>$\\
        Provide a brief justification here\\
      $<$/justification$>$\\
    $<$/verifiability$>$\\
    $<$visualization\_quality$>$\\
      $<$score$>$X$<$/score$>$\\
      $<$justification$>$\\
        Provide a brief justification here\\
      $<$/justification$>$\\
    $<$/visualization\_quality$>$\\
    $<$visualization\_consistency$>$\\
      $<$score$>$X$<$/score$>$\\
      $<$justification$>$\\
        Provide a brief justification here\\
      $<$/justification$>$\\
    $<$/visualization\_consistency$>$\\
  $<$report\_a$>$\\
  $<$report\_b$>$\\
    $<$!-- The same as above --$>$\\
  $<$report\_b$>$\\
$<$evaluation$>$\\
\\
\textbf{User prompt:}\\
\#\# Topic: \\
\{topic\}\\
\#\# learnings:\\
\{learnings\_str\}\\
$<$reportA$>$\\
    ...\\
    (base64 image into openai messages)\\
    ...\\
$<$/reportA$>$\\
$<$reportB$>$\\
    ...\\
    (base64 image into openai messages)\\
    ...\\
$<$/reportB$>$\\
\end{tcolorbox}

\subsection{Prompt for Chart Evaluation}
\label{appendix:prompts-chart-eval}
The prompt guides the evaluator to score the charts in the multimodal report based on five metrics: readability, layout, aesthetics, data faithfulness, and goal compliance.

\begin{tcolorbox}[breakable, colback=white, title={Prompt for Final Chart Assessment}]
\textbf{System prompt:}\\
You are an expert evaluator of AI-generated charts with advanced knowledge of data visualization and information analysis. Your role is to provide fair, impartial assessments of chart quality based strictly on objective criteria.\\
\\
\#\# Evaluation Task\\
You will evaluate a AI-generated chart based on:\\
- The chart itself\\
- The design specification used to generate the chart\\
\\
For each criterion below, assign a score from 1-10 (1=poor, 10=excellent). \\
\\
\#\# Evaluation Criteria\\
- Readability: Is the chart easy to read with appropriate titles, labels and colors?\\
- Layout: Is the layout of the chart appropriate with few or none issues such as overlapping?\\
- Aesthetics: Are the aesthetics of the visualization appropriate and effective for the visualization type and the data?\\
- Data Faithfulness: Is the data in the chart faithful to the data provided design specification?\\
- Goal compliance: How well the chart meets the specified visualization goals?\\
\\
\#\# Response Format:\\
Please give your response in the following XML format:\\
\\
$<$evaluation$>$\\
  $<$readability$>$\\
    $<$score$>$X$<$/score$>$\\
    $<$justification$>$\\
      Provide a brief justification here\\
    $<$/justification$>$\\
  $<$/readability$>$\\
  $<$layout$>$\\
    $<$score$>$X$<$/score$>$\\
    $<$justification$>$\\
      Provide a brief justification here\\
    $<$/justification$>$\\
  $<$/layout$>$\\
  $<$aesthetics$>$\\
    $<$score$>$X$<$/score$>$\\
    $<$justification$>$\\
      Provide a brief justification here\\
    $<$/justification$>$\\
  $<$/aesthetics$>$\\
  $<$data\_faithfulness$>$\\
    $<$score$>$X$<$/score$>$\\
    $<$justification$>$\\
      Provide a brief justification here\\
    $<$/justification$>$\\
  $<$/data\_faithfulness$>$\\
  $<$goal\_compliance$>$\\
    $<$score$>$X$<$/score$>$\\
    $<$justification$>$\\
      Provide a brief justification here\\
    $<$/justification$>$\\
  $<$/goal\_compliance$>$\\
$<$evaluation$>$\\
\textbf{User prompt:}\\
\{chart design and base64 image in report\}
\end{tcolorbox}

\section{Visualization examples}
\label{appendix:examples}

\subsection{Regular types of charts}
\begin{figure}[H]
    \centering
    \includegraphics[width=\linewidth]{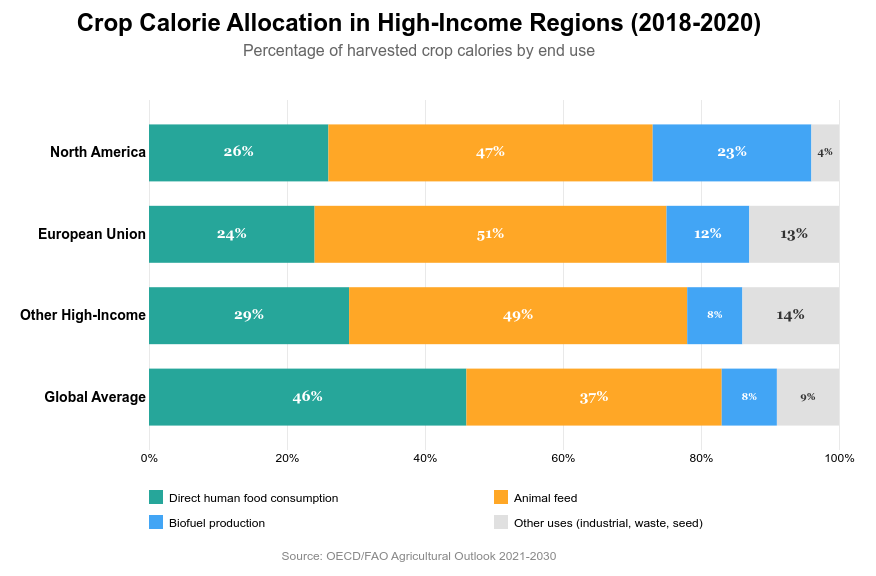}
    \caption{Example Bar Chart generated by Multimodal DeepResearcher}
    \label{fig:bar}
\end{figure}

\begin{figure}[H]
    \centering
    \includegraphics[width=\linewidth]{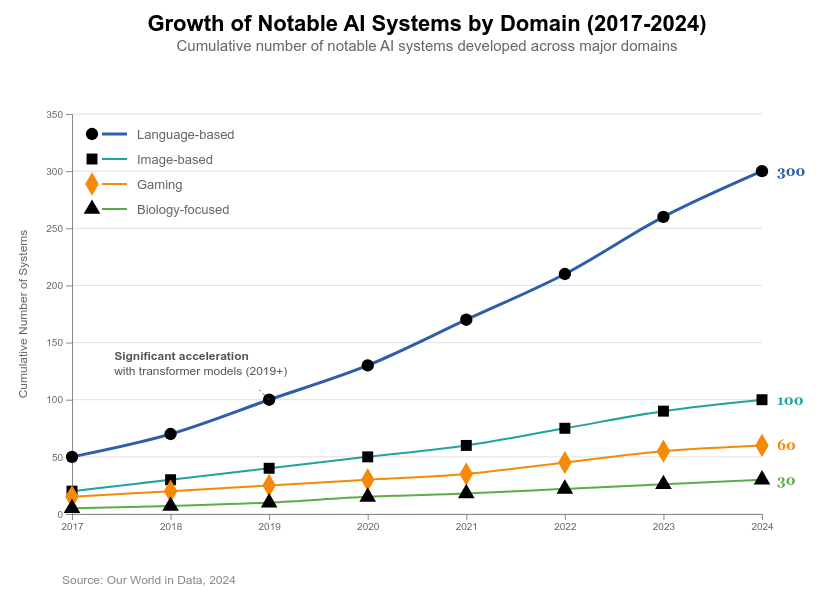}
    \caption{Example Line Chart generated by Multimodal DeepResearcher}
    \label{fig:line}
\end{figure}

\begin{figure}[H]
    \centering
    \includegraphics[width=\linewidth]{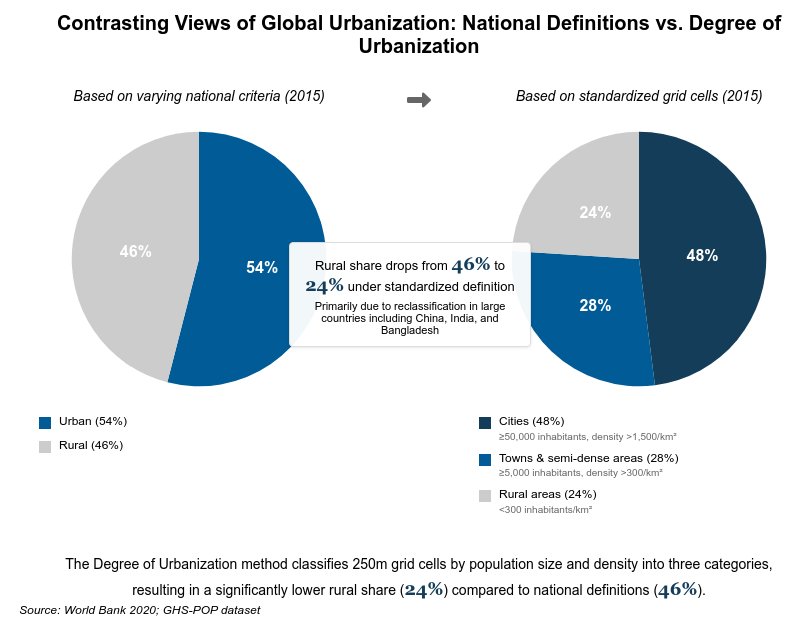}
    \caption{Example Pie Chart generated by Multimodal DeepResearcher}
    \label{fig:pie}
\end{figure}

\begin{figure}[H]
    \centering
    \includegraphics[width=\linewidth]{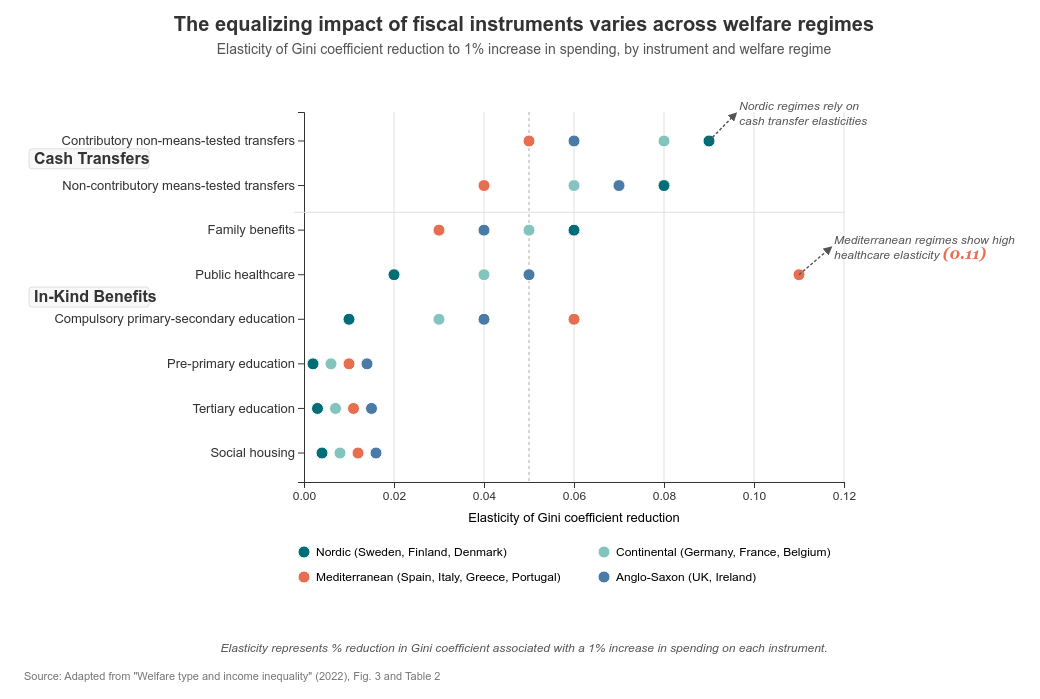}
    \caption{Example scatter chart generated by Multimodal DeepResearcher}
    \label{fig:scatter}
\end{figure}

\begin{figure}[H]
    \centering
    \includegraphics[width=\linewidth]{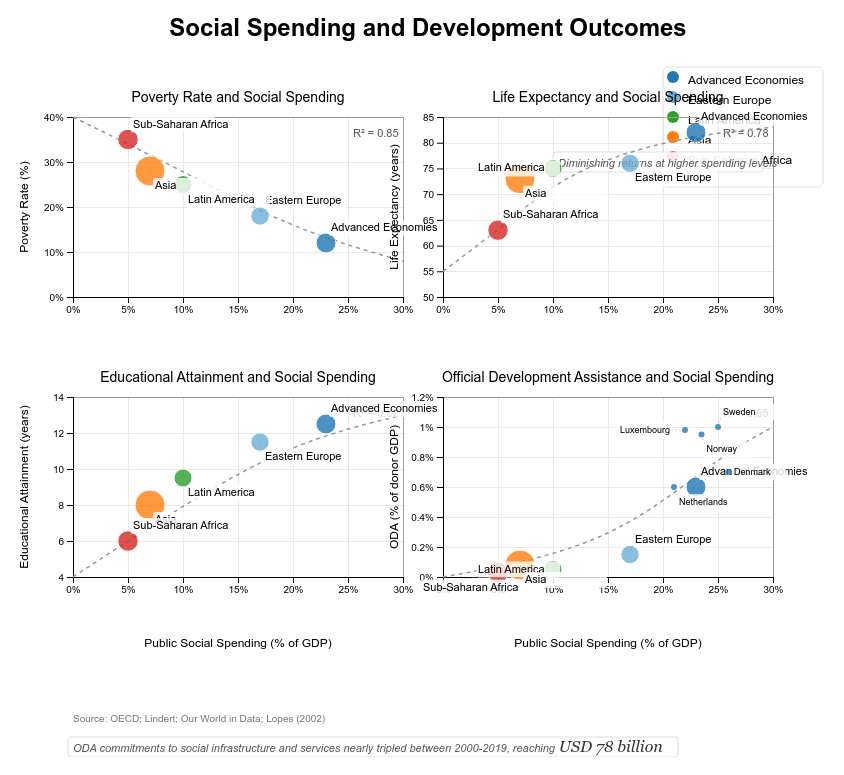}
    \caption{Example bubble chart generated by Multimodal DeepResearcher}
    \label{fig:bubble}
\end{figure}

\subsection{Sankey diagram}
\begin{figure}[H]
    \centering
    \includegraphics[width=\linewidth]{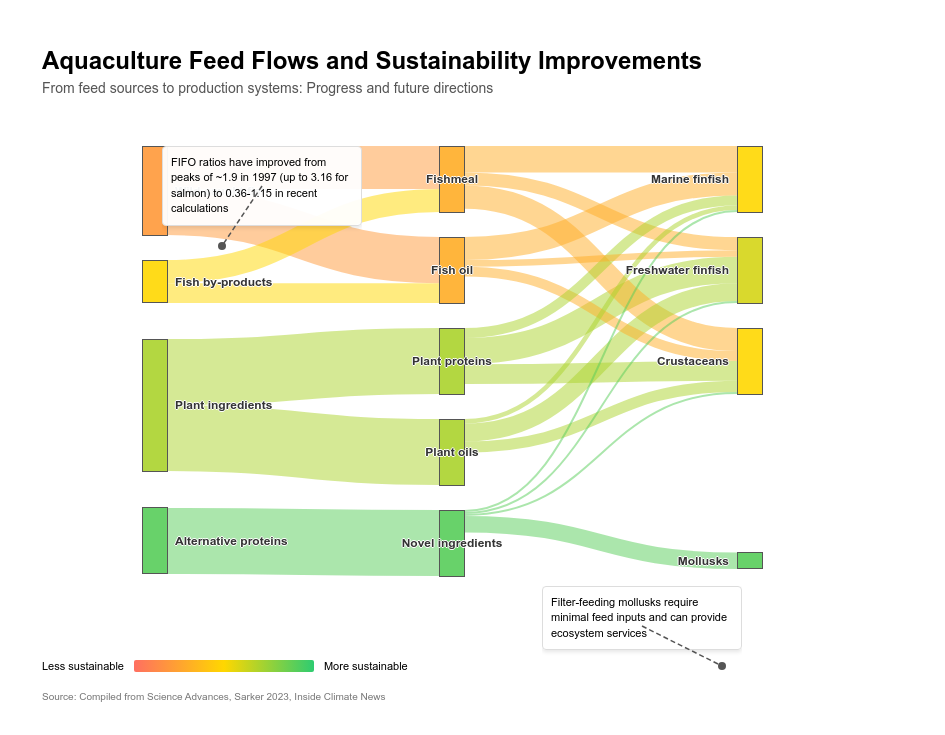}
    \caption{Example sankey diagram generated by Multimodal DeepResearcher}
    \label{fig:sankey}
\end{figure}

\subsection{Choropleth map}
\label{appendix:choropleth}
\begin{figure}[H]
    \centering
    \includegraphics[width=\linewidth]{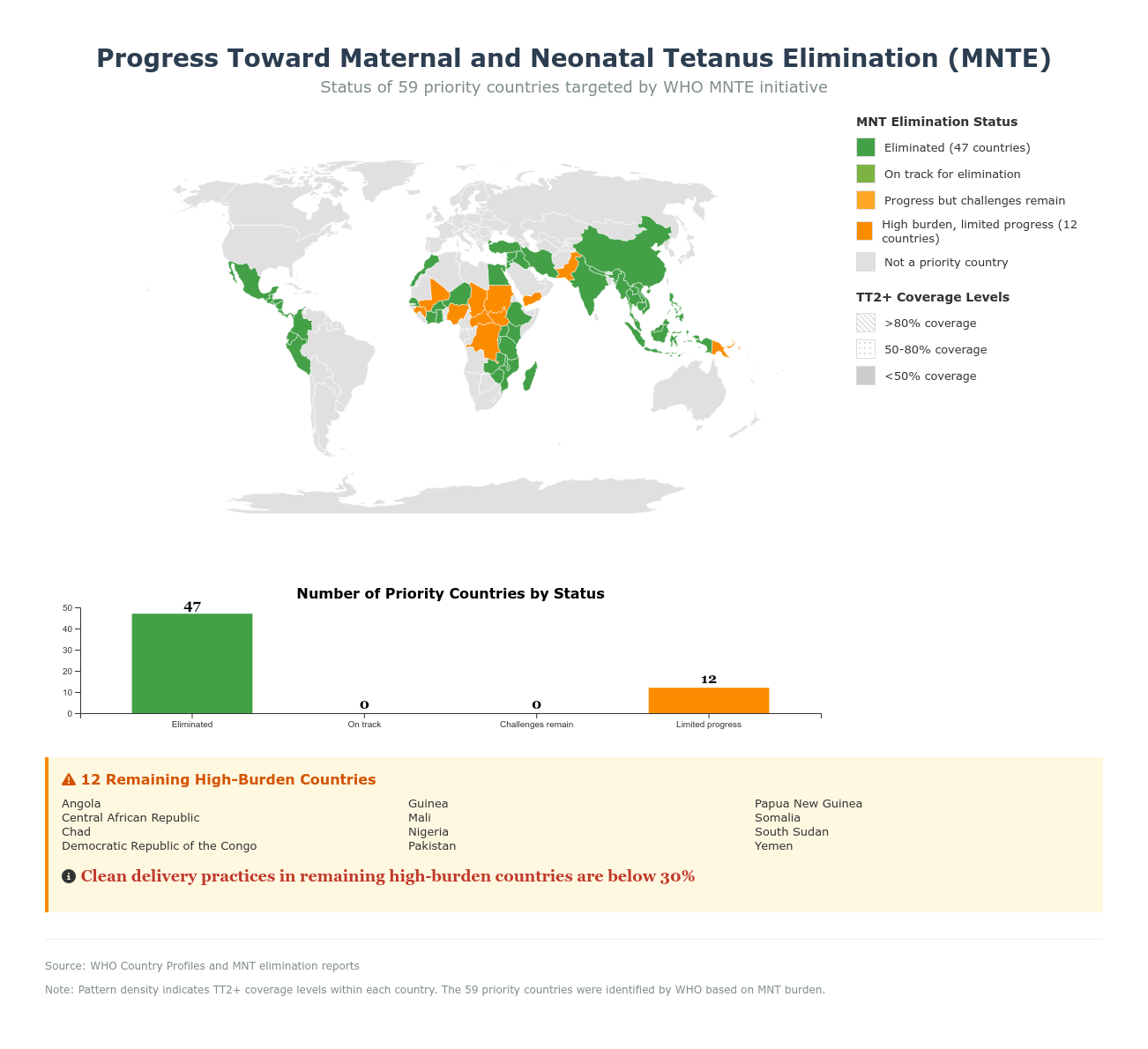}
    \caption{Example of Choropleth map generated by Multimodal DeepResearcher}
    \label{fig:map}
\end{figure}

\subsection{Flowchart}
\begin{figure}[H]
    \centering
    \includegraphics[width=\linewidth]{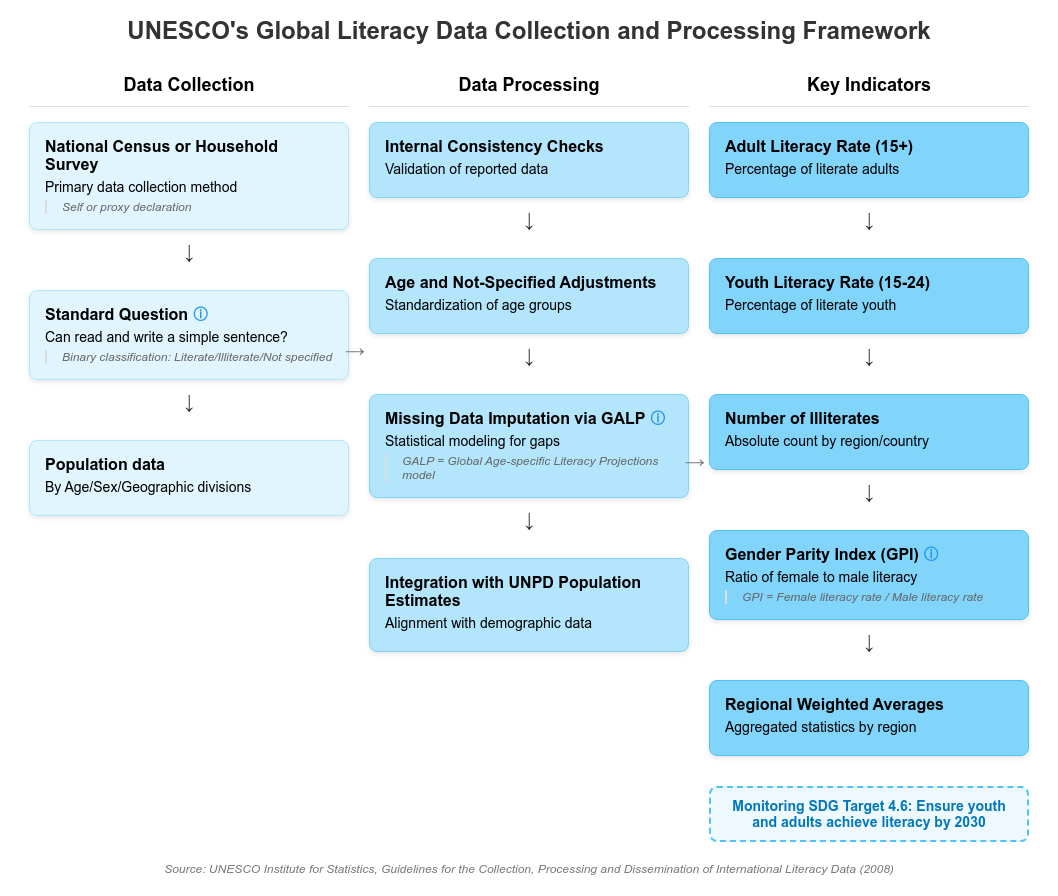}
    \caption{Example of flowchart generated by Multimodal DeepResearcher}
    \label{fig:flowchart}
\end{figure}

\subsection{Dashboard}
\begin{figure}[H]
    \centering
    \includegraphics[width=0.8\linewidth]{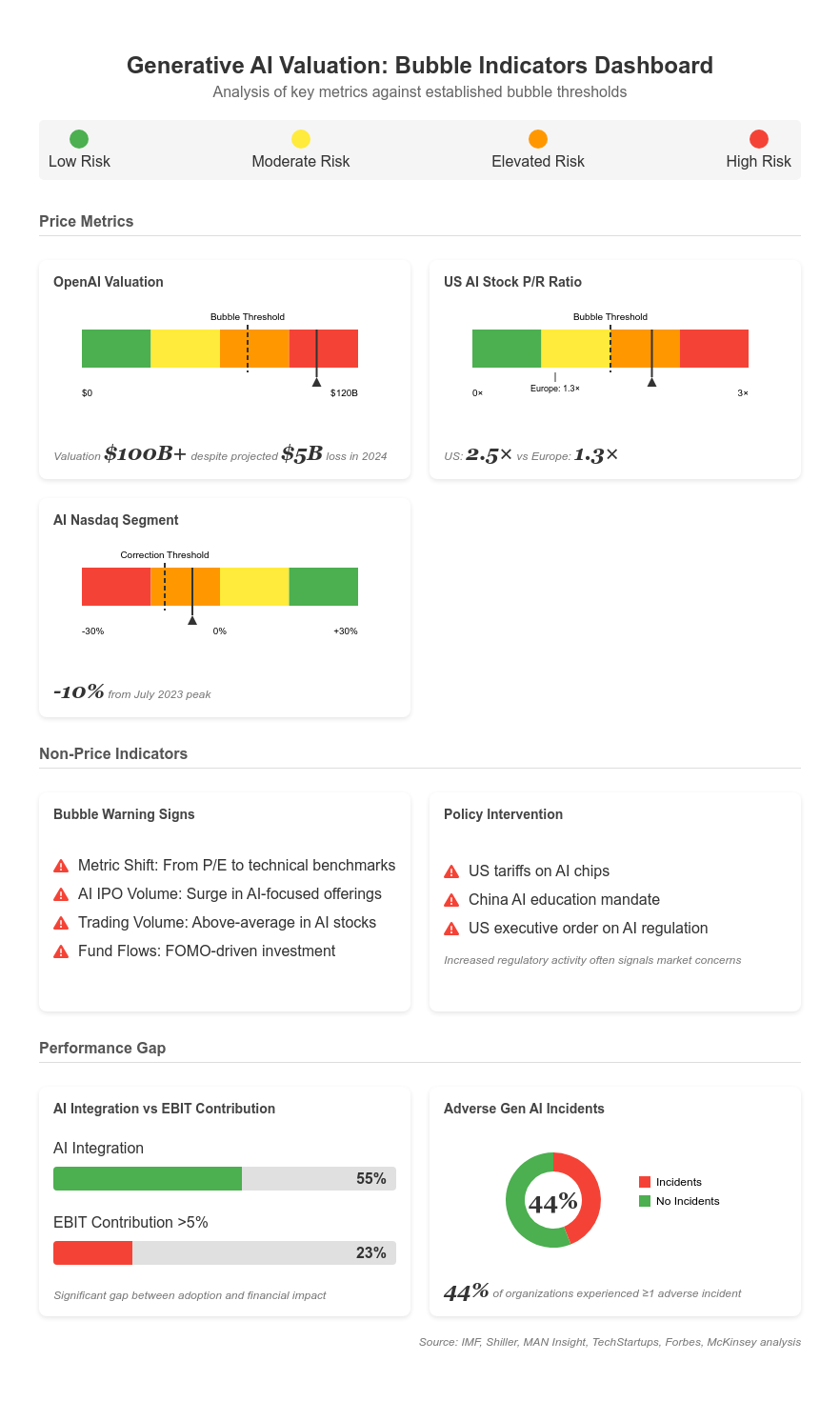}
    \caption{Example of dashboard map generated by Multimodal DeepResearcher}
    \label{fig:dashboard}
\end{figure}

\subsection{Infographic}
\begin{figure}[H]
    \centering
    \includegraphics[width=\linewidth]{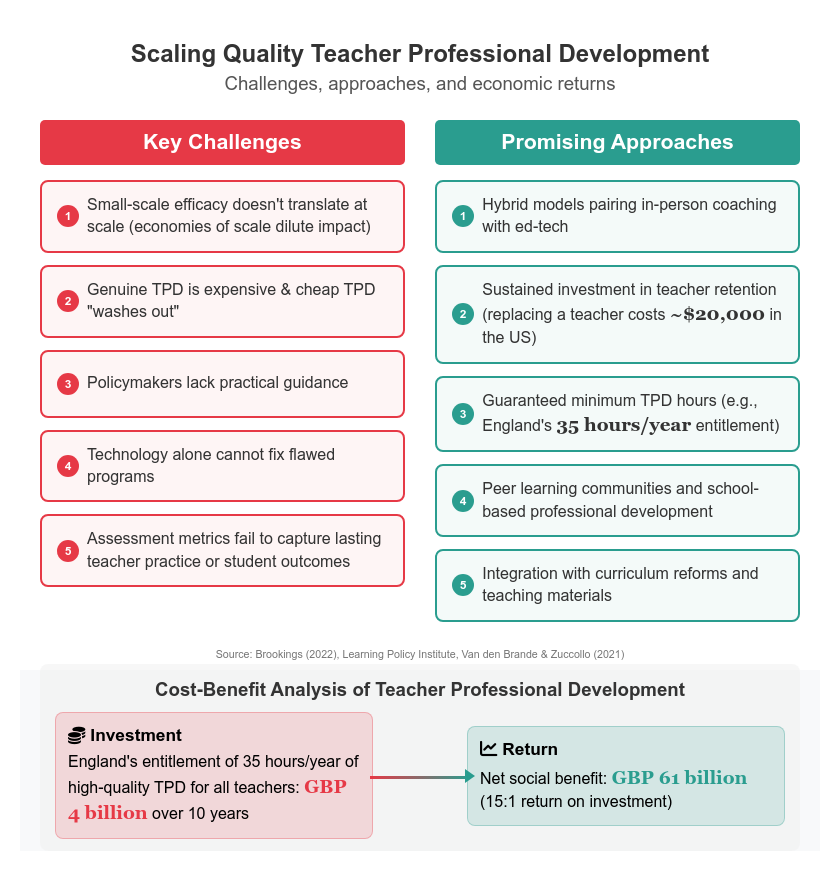}
    \caption{Example of infographic map generated by Multimodal DeepResearcher}
    \label{fig:infographic}
\end{figure}

\section{Error Case Examples}
\label{appendix:examples-error}
\begin{figure}[H]
    \centering
    \includegraphics[width=\linewidth]{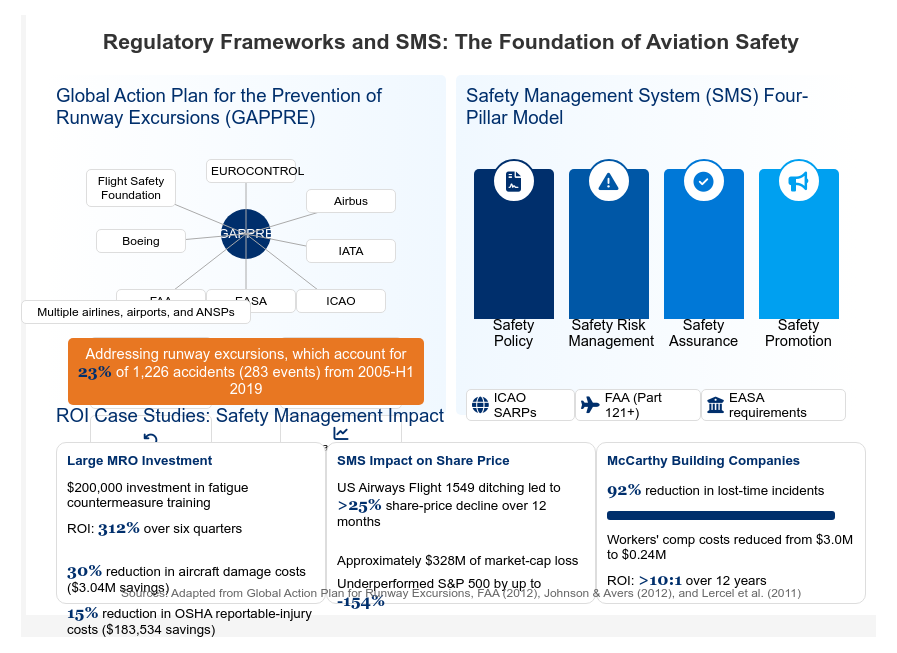}
    \caption{Overlapping caused by excessive information}
    \label{fig:overlapping1}
\end{figure}
\begin{figure}[H]
    \centering
    \includegraphics[width=\linewidth]{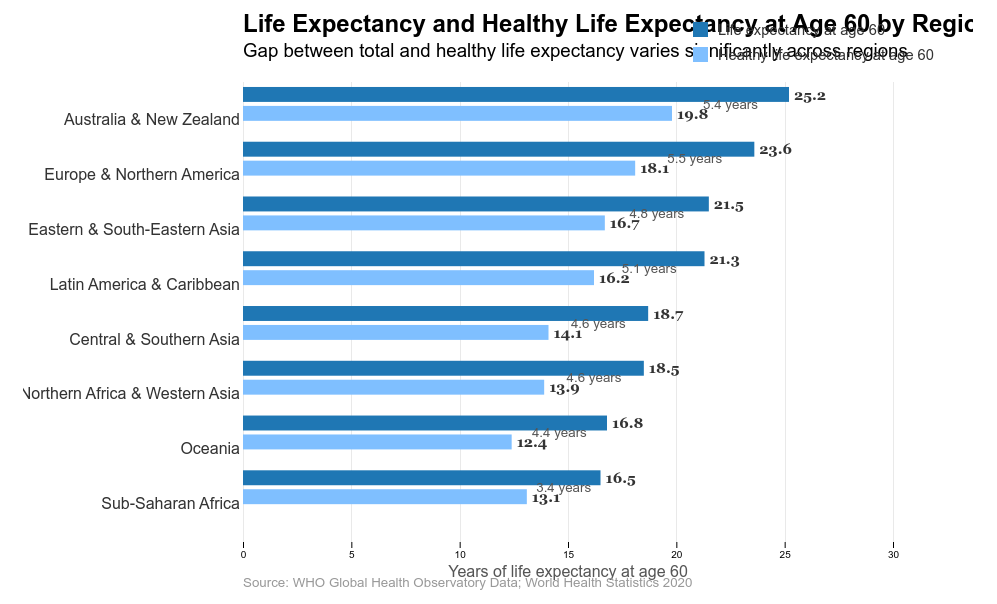}
    \caption{Overlapping caused by improper legend placement}
    \label{fig:overlapping2}
\end{figure}
\begin{figure}[H]
    \centering
    \includegraphics[width=\linewidth]{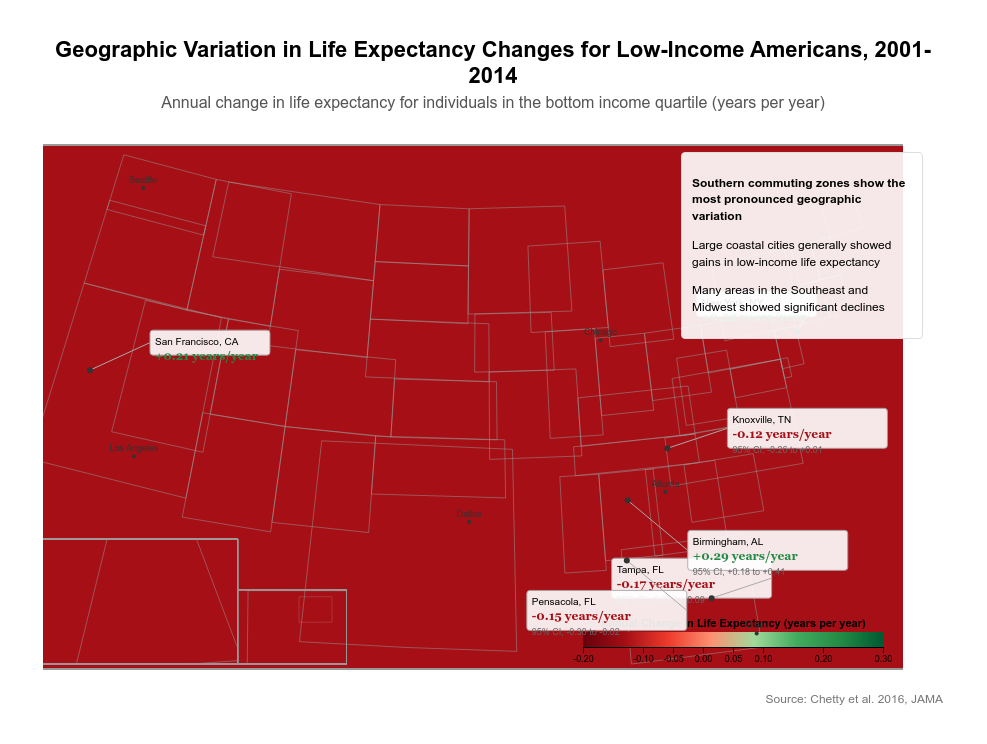}
    \caption{Hallucination in visualization generation}
    \label{fig:hallucination}
\end{figure}

\section{Full Report Example}
\label{appendix:example-report}
Below is a comprehensive report generated by the Multimodal DeepResearcher from scratch. The input topic is: \textit{Investments in waste management are key to ending plastic pollution}. For the sake of brevity, we have omitted the reference section of the report.



\section*{Supplementary material}

# Investments in Waste Management are Key to Ending Plastic Pollution

The global plastic pollution crisis represents one of the most urgent environmental challenges of our time. From riverine systems to the deep sea, plastic waste has penetrated virtually every ecosystem on Earth. While much attention focuses on reducing plastic production, the data reveals a critical insight: inadequate waste management infrastructure drives most plastic leakage into the environment, not production levels alone. This report examines the scale of the plastic pollution crisis, evaluates technical and policy solutions, identifies investment gaps, explores innovative financing mechanisms, and provides recommendations for strengthening global governance to address this pressing issue.

## Assessing the Plastic Pollution Crisis

The exponential growth of plastic production has created a waste management crisis of unprecedented scale. Since 1950, the world has produced approximately 9.2 billion tonnes of plastic, with production rates accelerating each decade. According to OECD's Global Plastics Outlook, global plastic waste generation more than doubled from 176 million tonnes in 2000 to 353 million tonnes in 2019.

Despite growing awareness of plastic pollution, current waste management practices remain woefully inadequate. Of all plastic waste generated globally:

- Only 9% is successfully recycled
- 50% ends up in landfills
- 19% is incinerated
- 22% is mismanaged

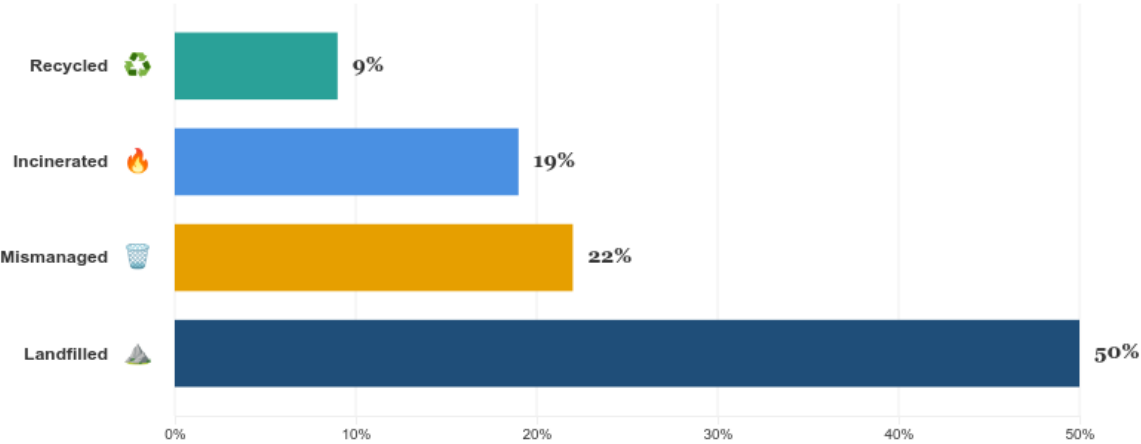

**Global Fate of Plastic Waste (2019)**

Despite growing awareness, **only 9%** of plastic waste is successfully recycled globally

Source: OECD Global Plastics Outlook (2022)

## Quantifying Plastic Leakage Pathways

The mismanagement of plastic waste has led to alarming levels of environmental leakage. In 2019 alone, approximately 6.1 million tonnes of plastic leaked into rivers and 1.7 million tonnes into oceans. This represents a significant portion of the 22% of global plastic waste that is mismanaged through practices such as open dumping, unsecured landfills, or uncontrolled burning.

Analyzing plastic leakage by region reveals substantial geographic disparities. Low- and middle-income countries bear a disproportionate burden of mismanaged waste despite consuming approximately three times less plastic per capita than high-income countries. According to Our World in Data, per-capita mismanaged plastic waste is highest in these regions, where rapid growth in plastic consumption has outpaced investments in waste management infrastructure.

A WWF-Dalberg report estimates the total lifetime cost of plastic at approximately $150 per kilogram in low- and middle-income countries—eight times the $19 per kilogram borne by high-income countries. This disparity grows even more extreme in low-income countries, where costs reach approximately $200 per kilogram, ten times that of high-income countries. These inequities stem from upstream design decisions made in high-income countries, under-resourced waste infrastructure in low- and middle-income countries, and lack of mandatory producer responsibility.

### The Inequitable Burden of Plastic Pollution

Despite lower consumption, LMICs bear higher economic and environmental costs

| High-Income Countries (HICs) | Low/Middle-Income Countries (LMICs) |
|---|---|

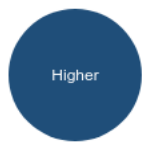

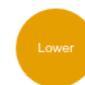

**Higher per-capita consumption**

🛒 Greater plastic use in everyday products

**$19 per kg**

$ Lifetime cost of plastic waste management

♻ Well-developed waste management systems

⚙ Design decisions made here

**3× less per-capita consumption**

🛍 Lower plastic use overall

**$150-200 per kg**

*8-10× higher than HICs*

$ Disproportionate economic burden

🗑 Under-resourced waste management

🌍 Bears environmental consequences

*Low-income countries face even higher costs, approximately $200 per kg (10× that of HICs)*

Source: WWF-Dalberg Report (November 2023)

## Country Case Studies

Country-specific examples further illustrate the scale of the challenge. In Brazil, for instance, over 10 million tonnes of plastic enter the domestic market annually, and the country imports 12,000 tonnes of plastic waste each year (growing at 7% annually). Yet only 22% of Brazilian municipalities collect waste for recycling. Based on current trajectories, Brazil could become the world's fourth-largest generator of mismanaged plastic waste.

The global cumulative stock of inadequately managed plastic waste has risen dramatically, from 61–72 million MT in 1990 to a projected 5,109–5,678 million MT by 2050, according to [Cordier et al.](). Interestingly, their regression analysis shows per-capita GDP growth explains just 11% of waste reduction, whereas extending average years of schooling could cut mismanaged waste by approximately 44% and improving corruption control by approximately 28%.

These findings underscore a critical insight: waste management infrastructure deficiencies—not merely plastic production volumes—drive plastic pollution. The data clearly demonstrates that addressing plastic pollution requires substantial investments in waste management systems, particularly in regions where infrastructure has not kept pace with consumption.

# Evaluating Waste Management & Policy Solutions

To effectively address plastic pollution, a comprehensive toolkit of technical and policy solutions is required. No single approach can solve the problem; rather, an integrated strategy tailored to local contexts offers the best path forward.

## Technical Solutions

The technical solutions available span the entire waste management chain:

1. **Collection and Segregation**: The foundation of effective waste management begins with reliable collection services and proper waste segregation. Without these fundamentals, downstream interventions have limited impact.

2. **Recycling Technologies**:

   - **Mechanical Recycling**: Processing plastics through sorting, cleaning, and reprocessing into new products.
   - **Chemical Recycling**: Breaking plastic polymers down into their chemical building blocks for reuse.

3. **Waste-to-Energy (WtE) Systems**: Converting non-recyclable plastics to energy through various thermal processes.

4. **Cleanup Technologies**: Specialized equipment for removing plastic from rivers, coastlines, and the ocean.

Recent cost-benefit analyses provide insights into the economic viability of these solutions. For instance, a study by Agori et al. (2024) in Ughelli, Nigeria, evaluated four mitigation strategies over a 0.5-year horizon at a 10% discount rate:

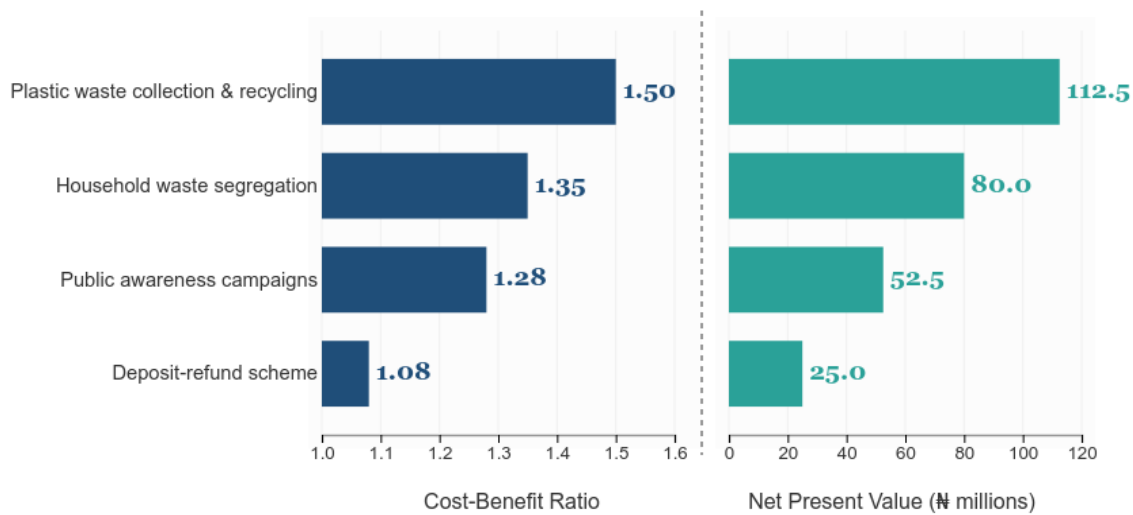

**Cost-Benefit Analysis of Plastic Waste Management Strategies in Ughelli, Nigeria**

Collection & recycling shows highest returns among evaluated interventions

Source: Agori et al. (2024). Evaluation over 0.5-year horizon at 10% discount rate.
Town population: 350,000; Plastic generation: 158,195 kg/year (0.452 kg/capita/year)

The results show that plastic waste collection and recycling delivered the highest returns with a Cost-Benefit Ratio (CBR) of 1.50 and Net Present Value (NPV) of ₦112,500,000, followed by household waste segregation (CBR 1.35, NPV ₦80,000,000), public awareness campaigns (CBR 1.28, NPV ₦52,500,000), and deposit-refund schemes (CBR 1.08, NPV ₦25,000,000).

In terms of waste-to-energy systems, Khwammana & Chaiyata (2025) reported on a waste-to-energy-to-zero system that uses municipal solid waste (17.85 tonnes/day at 31.63% combustible) to fuel a combined cooling, heating, and power plant. The system delivers 306.98 kW at 22.38% efficiency, yielding a levelized energy cost of 0.15 USD/kWh, NPV of 1,634,658 USD, profitability index of 1.72, internal rate of return of 7.97%, and payback period of 9.63 years.

Clement's (2012) Fort Bliss WtE/CSP hybrid cost-benefit study shows that NPV is highly sensitive to the gap between local tariff and WtE rate. Using EPA's WARM model for 1 million tonnes/year feedstock, the Fort Bliss WtE diversion avoids approximately 264,025 MTCO2e annually. At carbon credit prices ranging from 0.10-10 USD/MTCO2e, 20-year environmental benefits range from $0.4 million to $36.2 million USD.

## Policy Instruments

While technical solutions provide the means to manage plastic waste, policy instruments create the enabling environment and incentives necessary for implementation:

1. **Extended Producer Responsibility (EPR)**: Schemes that make manufacturers responsible for the entire lifecycle of their products, including end-of-life treatment.

2. **Deposit-Return Systems (DRS)**: Programs that incentivize consumers to return used packaging for recycling.

3. **Economic Instruments**:

   - **Pollution Taxes**: Levies on plastic products or packaging.
   - **Performance Bonds**: Financial guarantees required from producers to ensure proper waste management.
   - **Carbon Pricing**: Mechanisms that assign a cost to carbon emissions, relevant for plastic production and waste management.

4. **Regulatory Measures**:

   - **Bans and Restrictions**: Prohibitions on specific single-use plastic items.

- **Recycled Content Mandates**: Requirements for minimum percentages of recycled material in products.

The effectiveness of these policy instruments varies by context. For instance, in Small Island Developing States (SIDS), which contribute only 1.64% of global plastic waste generation but receive over 93% of macroplastic leakage, policy responses include bans, levies (0.12-0.40% of tourism revenue), and deposit-refund schemes (such as Palau's US $30 arrival fee that funds plastic buy-back).

## Comparative Cost-Effectiveness

A critical finding from recent research is that no single solution exists for plastic pollution. Technical options such as mechanical/chemical recycling, incineration, river and ocean cleanup devices, and wastewater treatments yield highly variable CAPEX/OPEX profiles. Similarly, policy measures (bans, levies, EPR, informal-sector support, voluntary targets) must be tailored to local contexts.

**Cost-Effectiveness of Plastic Pollution Solutions by Country Income Level**

*No single solution works in all contexts—integrated approaches perform best*

|  | Low-Income | Middle-Income | High-Income |
| --- | --- | --- | --- |
| **Technical Solutions** |  |  |  |
| Collection & segregation systems | Medium | High | High |
| Mechanical recycling | Medium | High | High |
| Chemical recycling |  | Medium | High |
| Waste-to-energy |  | Medium | High |
| River cleanup technologies | Medium | Medium | Low |
| Ocean cleanup technologies | Low | Low | Low |
| **Policy Instruments** |  |  |  |
| Extended Producer Responsibility | Low | Medium | High |
| Deposit-return schemes | Low | Medium | Medium |
| Single-use plastic bans | High | High | Medium |
| Pollution taxes | Low | High | High |
| Informal sector support | High | Medium | Medium |
| Public awareness campaigns | Medium | Medium | Medium |

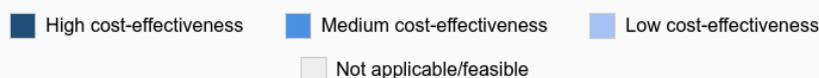

■ High cost-effectiveness   ■ Medium cost-effectiveness   ■ Low cost-effectiveness
□ Not applicable/feasible

Source: Based on cost-effectiveness review data (2021)

As shown in the visualization, cost-effectiveness varies substantially by country income level:

- In low-income countries, collection and segregation systems, informal sector support, and public awareness campaigns yield the highest cost-effectiveness.

- In middle-income countries, mechanical recycling and extended producer responsibility schemes show stronger returns.
- In high-income countries, advanced technical solutions like chemical recycling and comprehensive policy frameworks deliver the best results.

This framework highlights the importance of adopting integrated approaches tailored to local economic contexts to maximize tonnage reduced per dollar spent.

# Mapping Investment Needs and Financing Gaps

The scale of plastic pollution requires substantial financial resources to implement effective waste management solutions. Under a business-as-usual scenario, mismanaged plastics waste is projected to exceed 250 million metric tons annually by 2040 and will require an estimated USD 1.64 trillion in redirection of financial flows toward circularity solutions, according to UNEP's Turning off the Tap report.

## Current Financing Landscape

Despite the magnitude of the challenge, current financing for plastic waste management remains woefully inadequate. Official development finance (ODF) for solid waste management is minimal:

- Only **US $1.8 billion** in ODF commitments in 2021
- This represents just **0.41%** of all ODF
- Low-income countries captured only **8%** of that amount

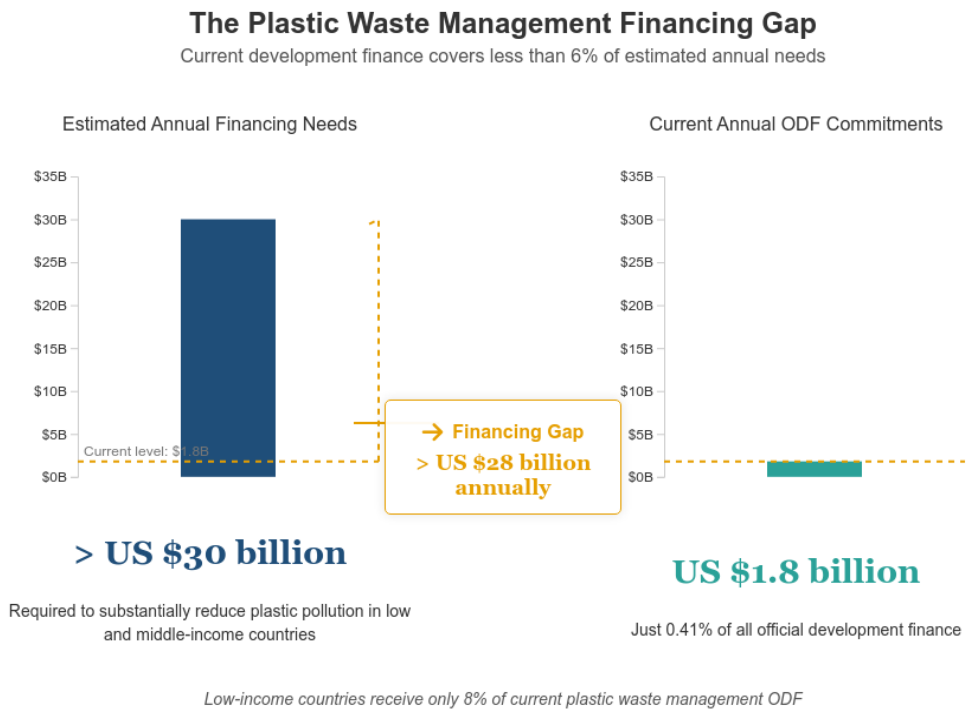

This stark contrast between estimated needs (>US $30 billion annually) and current commitments (US $1.8 billion) creates a financing gap of more than US $28 billion per year. This gap is particularly concerning given that low-income countries, where the problem is most acute, receive only a small fraction of available funding.

## Lessons from Climate Finance

The challenges in mobilizing adequate funding for plastic waste management parallel those faced in climate finance. However, there are encouraging lessons to be drawn. Global climate finance flows doubled from USD 364 billion in 2011 to USD 850 billion in 2020, with private finance constituting approximately 50 percent of the total.

The growth rates differ significantly between public and private sources:

- Public finance grew at 9.1 percent annually
- Private finance grew at only 4 percent annually

This highlights the critical role of public finance in catalyzing and leading investment growth, while also pointing to the untapped potential of private finance if appropriate mechanisms are developed.

## Regional and Sector Distribution

The financing needs are not evenly distributed across regions or sectors. Low- and middle-income countries require the bulk of investment due to their infrastructure deficits. Similarly, certain sectors within the waste management value chain—particularly collection, sorting, and recycling infrastructure—require more significant capital infusion.

The Global Environment Facility (GEF) has recognized this need and scaled its plastic-pollution financing from US $10 million in GEF-6 (2014-18) to US $840 million in GEF-8 (2022-26), leveraging US $5 billion in co-financing and preventing nearly 25 million tonnes of plastics from entering waste streams.

Despite these positive developments, the current allocation of funding is heavily skewed toward downstream waste management (80% of available funding) rather than upstream prevention (only 3%). This imbalance needs to be addressed to ensure a comprehensive approach to the plastic pollution challenge.

# Deploying Innovative Finance Mechanisms

Addressing the substantial financing gap for plastic waste management requires creative financial solutions that can mobilize both public and private capital. Blended finance—combining different types of capital with varying risk-return profiles—offers a promising approach to unlock investments in circular plastics.

## Landmark Blended Finance Transactions

Several groundbreaking transactions demonstrate the potential of blended finance to catalyze investments in circular plastics:

# Landmark Blended Finance Transactions in Circular Plastics

Case studies of successful capital mobilization for waste management and recycling

## Indorama Ventures

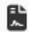 Blue Loan

**US $300 million** (IFC $150M; ADB $100M; DEG $50M)

Underwriting **US $1.5 billion** rPET capacity across 5 emerging markets

**US $20 million** JV plant in Philippines (30 ktpa, 2 billion bottles)

## Belize Blue Bond

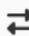 Sovereign Debt Conversion

**US $364 million** Blue Bond structured by The Nature Conservancy and Credit Suisse

Repurchased US $546 million Superbond at **45% discount**

Commitment to expand marine protected areas from 16% to **30%** of territorial seas by 2026

## Mexico's ECOCE (2002) and India's EPR Framework (2022)

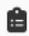 Policy-Enabled Investment

**US $26 million** funding for Recykal's digital EPR platform

**US $339 million** invested in 16 PET plants in Mexico

Raised Mexico's PET recycling rate from 8.8% (2002) to **56%** (2018)

Blended finance combines public, philanthropic, and private capital with varying risk-return profiles to mobilize investment in sustainable development.

1. **Indorama Ventures**: Secured the world's first non-sovereign US $300 million blue loan (IFC $150 million; ADB $100 million; DEG $50 million) to underwrite US $1.5 billion in recycled PET (rPET) capacity across five emerging markets. The company also established a US $20 million joint venture plant in the Philippines with a capacity of 30,000 tonnes per annum, equivalent to recycling 2 billion bottles.
2. **Belize Blue Bond**: The Nature Conservancy and Credit Suisse structured a US $364 million Blue Bond—backed by DFC political risk insurance—to repurchase a US $546 million Superbond at a 45% discount. As part of this transaction, Belize committed to expanding its marine protected areas from 16% to 30% of territorial seas by 2026.
3. **Policy-Enabled Investments**: Extended Producer Responsibility (EPR) frameworks in Mexico (ECOCE, established in 2002) and India (implemented in 2022) have catalyzed significant investments. These include US $26 million in funding for Recykal's digital EPR platform and

US $339 million invested in 16 PET recycling plants in Mexico, helping to raise Mexico's PET recycling rate from 8.8% in 2002 to 56% in 2018.

## Sustainability-Linked Financial Instruments

Beyond traditional blended finance structures, sustainability-linked financial instruments are emerging as powerful tools for mobilizing capital:

1. **Sustainability-Linked Bonds (SLBs)**: Natura's US $1 billion SLB (May 2021) ties a 0.65% coupon step-up to achieving 25% post-consumer recycled plastic use (from 9% in 2019) and 13% greenhouse gas emissions reductions by 2026.
2. **Green Sukuk**: Indonesia has issued green sukuk (Islamic bonds) totaling US $1.25 billion (2018) and a US $750 million green tranche (2021), with up to 17% of proceeds allocated to waste management.
3. **Blue Bonds**: Similar to green bonds but specifically focused on marine and ocean conservation, blue bonds can channel capital toward projects that reduce plastic pollution in marine environments.

## Overcoming the "Missing Middle" Financing Gap

A particular challenge in financing plastic waste management is addressing the "missing middle"—small and medium-sized enterprises (SMEs) that are too large for microfinance but too small for conventional financing. This gap is especially pronounced in low- and middle-income countries, where many waste management operations are SMEs.

To overcome this challenge, several innovative approaches can be deployed:

1. **SME Aggregation**: Pooling multiple small projects into larger investment vehicles to achieve scale and reduce transaction costs.
2. **Partial Credit Guarantees**: Risk-sharing mechanisms that encourage lenders to extend credit to SMEs by providing partial coverage of potential losses.
3. **First-Loss Facilities**: Financial structures that absorb initial losses on a portfolio of investments, reducing risk for other investors.
4. **Technical Assistance Facilities**: Support services to improve the bankability of projects through business development, operational improvements, and capacity building.

For instance, concessional facilities administered by public finance institutions and multilateral development banks can pool sub-USD 2 million circularity projects into asset-backed vehicles secured by partial credit guarantees. This approach reduces financing costs, addresses SMEs' limited ticket size, and "crowds in" institutional investors.

# Strengthening Governance & Global Finance Architecture

Effective waste management and plastic pollution reduction require not only financial resources but also robust governance frameworks at both national and international levels. The evolving global landscape presents opportunities to strengthen governance and create a more supportive environment for investment.

# International Plastics Treaty and Financial Mechanisms

The development of an international legally binding instrument (ILBI) on plastic pollution represents a significant opportunity to establish a coordinated global response. The Zero Draft of the future ILBI proposes a "Mechanism...including financial resources from all sources, domestic and international, public and private," creating the vehicle to embed blended finance tools—blue loans, concessional lending, partial credit guarantees, first-loss facilities—to de-risk plastics circularity projects and mobilize private capital.

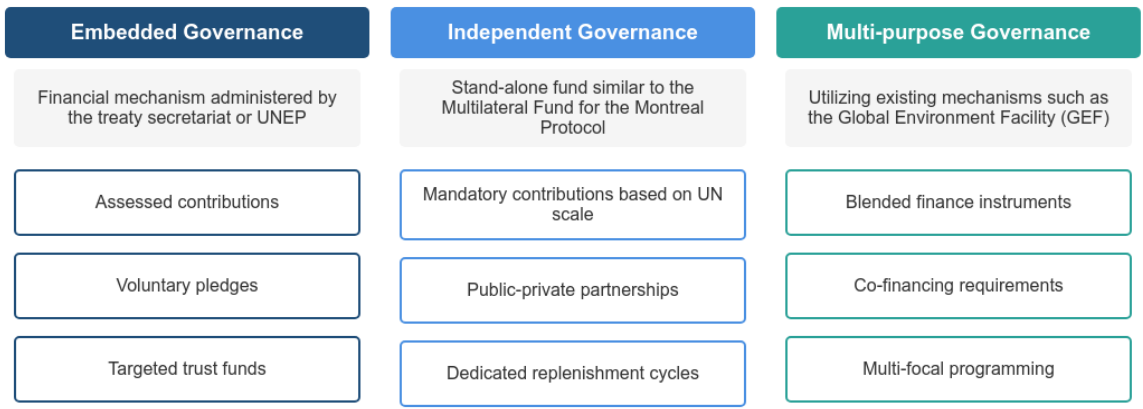

**Potential Financial Mechanisms Under an International Plastics Treaty**
Governance options and funding instruments to support implementation

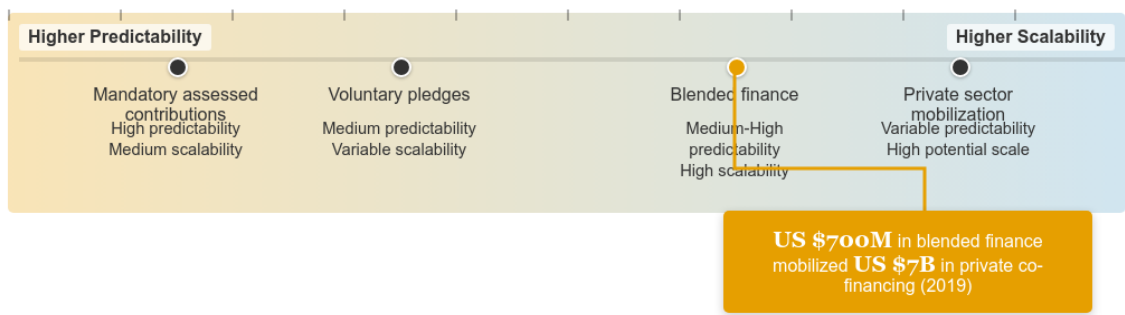

**Contribution Approaches: Predictability and Scalability**

Source: UNEP (2005, 2020b); GEF Reports

Three primary governance options are being considered for financial mechanisms under a plastics treaty:

1. **Embedded Governance**: Financial mechanism administered by the treaty secretariat or UNEP.
2. **Independent Governance**: Stand-alone fund similar to the Multilateral Fund for the Montreal Protocol.
3. **Multi-purpose Governance**: Utilizing existing mechanisms such as the Global Environment Facility (GEF).

Each approach has advantages and limitations regarding funding predictability, administrative efficiency, and ability to mobilize private capital. A hybrid approach may ultimately be most effective.

## Finance Needs Across Three Action Areas

Key finance needs in a legally binding plastics treaty cover three action areas:

1. **Enabling Activities**: National inventories, policy design, and enforcement capacity. Currently, only 3% of available funding under AHEG's inventory serves upstream prevention, versus 80% for downstream waste management.
2. **Knowledge-Related Activities**: Clearing-house mechanisms, global science-policy interface, and R&D for sustainable plastic alternatives.
3. **Circular Economy Transitions**: Innovative product/process design, collection/recycling infrastructure, market development for recycled resins.

## Harmonizing Financial Standards and Disclosures

Mandatory harmonization of sustainable finance taxonomies and metrics, coupled with mandatory corporate disclosures on plastics-related risks, dependencies, and impacts (as proposed in Zero Draft Part II.13), can lower investor risk premiums. This transparency and policy certainty are prerequisites for structuring concessional "blue loans" targeting marine plastic cleanup.

Strong Extended Producer Responsibility (EPR) schemes, such as Malaysia's mandatory EPR by 2026 under the Malaysia Plastics Sustainability Roadmap 2021-2030, provide a supportive policy environment for investments in waste management infrastructure.

## Special Considerations for Vulnerable Regions

Small Island Developing States (SIDS) represent a special case that requires targeted support. SIDS contribute only 1.64% of global plastic waste generation and 1.56% of mismanaged plastic waste but receive over 93% of macroplastic leakage. These nations struggle with geography-driven high waste-management costs, poor data, limited enforcement capacity, and lack of financing (local budgets cover only about 20% of municipal spending on waste).

Regional programs such as the Pacific Regional Action Plan, developed under the Secretariat of the Pacific Regional Environment Programme (SPREP), provide coordinated approaches to address plastic pollution in these vulnerable regions. However, SIDS need treaty support for capacity building, technology transfer, and dedicated official development assistance.

# Conclusion and Recommendations

The plastic pollution crisis presents a formidable challenge that requires urgent action and substantial financial resources. While the focus often centers on reducing plastic production, the data clearly shows that inadequate waste management infrastructure is the primary driver of plastic leakage into the environment.

Key findings from this analysis include:

1. **Scale of the Challenge**: With global plastic waste generation exceeding 353 million tonnes annually and only 9% successfully recycled, the magnitude of the problem is immense.
2. **Infrastructure Gap**: Low- and middle-income countries bear a disproportionate burden of the impacts despite consuming less plastic per capita, primarily due to insufficient waste management infrastructure.
3. **Financial Shortfall**: Current official development finance for waste management (US $1.8 billion annually) falls far short of the estimated US $30+ billion needed to substantially reduce plastic pollution.

4. **Integrated Solutions**: No single solution is universally effective; instead, a tailored approach combining technical and policy interventions based on local contexts yields the best results.
5. **Innovative Finance**: Blended finance mechanisms and sustainability-linked instruments show promise in mobilizing private capital for waste management infrastructure.

To address these challenges effectively, we recommend the following actions:

## For Policymakers:

1. Implement robust Extended Producer Responsibility (EPR) schemes that create economic incentives for better product design and end-of-life management.
2. Develop and enforce regulatory standards for waste collection, recycling, and disposal.
3. Harmonize sustainable finance taxonomies and require corporate disclosure of plastic-related impacts.
4. Ensure that international agreements include adequate financial mechanisms with both mandatory and voluntary contributions.

## For Financial Institutions:

1. Scale up blended finance initiatives that can de-risk investments in waste management infrastructure in emerging markets.
2. Develop innovative financial products such as blue bonds, sustainability-linked loans, and green sukuk specifically targeting plastic pollution reduction.
3. Create aggregation mechanisms to address the "missing middle" financing gap for SMEs in the waste management sector.

## For Waste Management Operators:

1. Implement cost-effective collection and segregation systems as foundations for successful waste management.
2. Explore partnerships with international financial institutions and impact investors to access capital for infrastructure development.
3. Adopt technologies and business models appropriate to local conditions and available resources.

## For the International Community:

1. Finalize a robust international legally binding instrument on plastic pollution with clear targets, standards, and financial support mechanisms.
2. Increase official development finance for waste management, particularly targeting low-income countries where needs are greatest.
3. Support knowledge sharing and capacity building to ensure effective implementation of technical and policy solutions.

Investments in waste management infrastructure represent the most direct and effective pathway to reducing plastic pollution in the environment. By mobilizing adequate financial resources, implementing appropriate technologies, and creating supportive policy frameworks, we can make significant progress toward ending this global environmental crisis.

The time for action is now. Each year of delay adds millions of tonnes of plastic to our oceans, rivers, and landscapes. With concerted effort and strategic investments, we can build a more sustainable future where plastics no longer pose a threat to our environment and well-being.


% appendix end

\end{document}